\documentclass[10pt,twocolumn,letterpaper]{article}

\usepackage{iccv}              % To produce the CAMERA-READY version
\usepackage{balance}
\definecolor{iccvblue}{rgb}{0.21,0.49,0.74}
\usepackage[pagebackref,breaklinks,colorlinks,allcolors=iccvblue]{hyperref}
\usepackage{multirow}

\def\paperID{10327} % *** Enter the Paper ID here
\def\confName{ICCV}
\def\confYear{2025}

\title{Adver-City: Open-Source Multi-Modal Dataset for Collaborative Perception Under Adverse Weather Conditions}

\author{Mateus Karvat$^{1}$ and Sidney Givigi$^{1}$\\
$^{1}$School of Computing, Queen's University, Canada\\
{\tt\small \{mateus.karvat, sidney.givigi\}@queensu.ca}
}

\begin{document}
\maketitle
\begin{abstract}
Adverse weather conditions pose a significant challenge to the widespread adoption of Autonomous Vehicles (AVs) by impacting sensors like LiDARs and cameras. Even though Collaborative Perception (CP) improves AV perception in difficult conditions, existing CP datasets lack adverse weather conditions. To address this, we introduce Adver-City, the first open-source synthetic CP dataset focused on adverse weather conditions. Simulated in CARLA with OpenCDA, it contains over $24$ thousand frames, over $890$ thousand annotations, and $110$ unique scenarios across six different weather conditions: clear weather, soft rain, heavy rain, fog, foggy heavy rain and, for the first time in a synthetic CP dataset, glare. It has six object categories including pedestrians and cyclists, and uses data from vehicles and roadside units featuring LiDARs, RGB and semantic segmentation cameras, GNSS, and IMUs. Its scenarios, based on real crash reports, depict the most relevant road configurations for adverse weather and poor visibility conditions, varying in object density, with both dense and sparse scenes, allowing for novel testing conditions of CP models. Benchmarks run on the dataset show that weather conditions created challenging conditions for perception models, with CoBEVT scoring $58.30/52.44/38.90$ (AP@30/50/70). The dataset, code and documentation are available at \url{https://labs.cs.queensu.ca/quarrg/datasets/adver-city/}.
\end{abstract}    
\section{Introduction}
\label{sec:intro}

Self-driving cars are here, with robotaxi companies in the USA and China having already logged over $60$ million fully autonomous miles on public roads~\cite{bishop_2024_million}. However, Autonomous Vehicles (AVs) still have a long road ahead before reaching widespread adoption. Many issues must still be addressed to ensure that AVs can drive safely and reliably in all possible scenarios.

Adverse weather conditions such as rain, fog, and glare present significant challenges to AVs. Although most driving is not done in these conditions, they create particularly difficult driving scenarios. For instance, it only rains on land $8\%$ of the time \cite{trenberth_2018_how} but the risk of accidents in rainy weather is $70\%$ higher than normal~\cite{andrey_1993_temporal}. Furthermore, adverse weather degrades the performance of perception sensors such as cameras and LiDARs~\cite{ozarkar_2022_physicsbased}. 

A promising solution to this challenge is Collaborative Perception (CP)~\cite{zhang_2023_perception}, a growing field within self-driving research. In CP, Connected AVs (CAVs) share information about their surroundings with each other, an approach called V2V (Vehicle-to-Vehicle) collaboration, or also with other entities such as Roadside Units (RSUs), an approach called V2X (Vehicle-to-Everything) collaboration. CP significantly enhances a vehicle's perception capabilities~\cite{han_2023_collaborative}, especially in situations involving occlusion or long-range detection~\cite{wang_2023_deepaccident}, and is expected to offer similar benefits in adverse weather conditions.

However, the development of robust CP models for challenging weather is hindered by the fact that no public real-world CP dataset focuses on adverse weather conditions. Despite their reduced realism, simulated datasets offer a promising solution to such a shortage through controlled scenario design, allowing uncommon scenarios such as heavy fog and intense glare to be captured extensively. Yet, existing open-source CP datasets have no adverse weather scenes, while those with adverse conditions offer little weather diversity and are not open-source, preventing scene modifications to expand weather variety.

\begin{figure*}
\centering
    \begingroup
    \hspace*{-0.25cm}
    \begin{tabular}{c@{}c@{}c}
        \includegraphics[width=.325\textwidth]{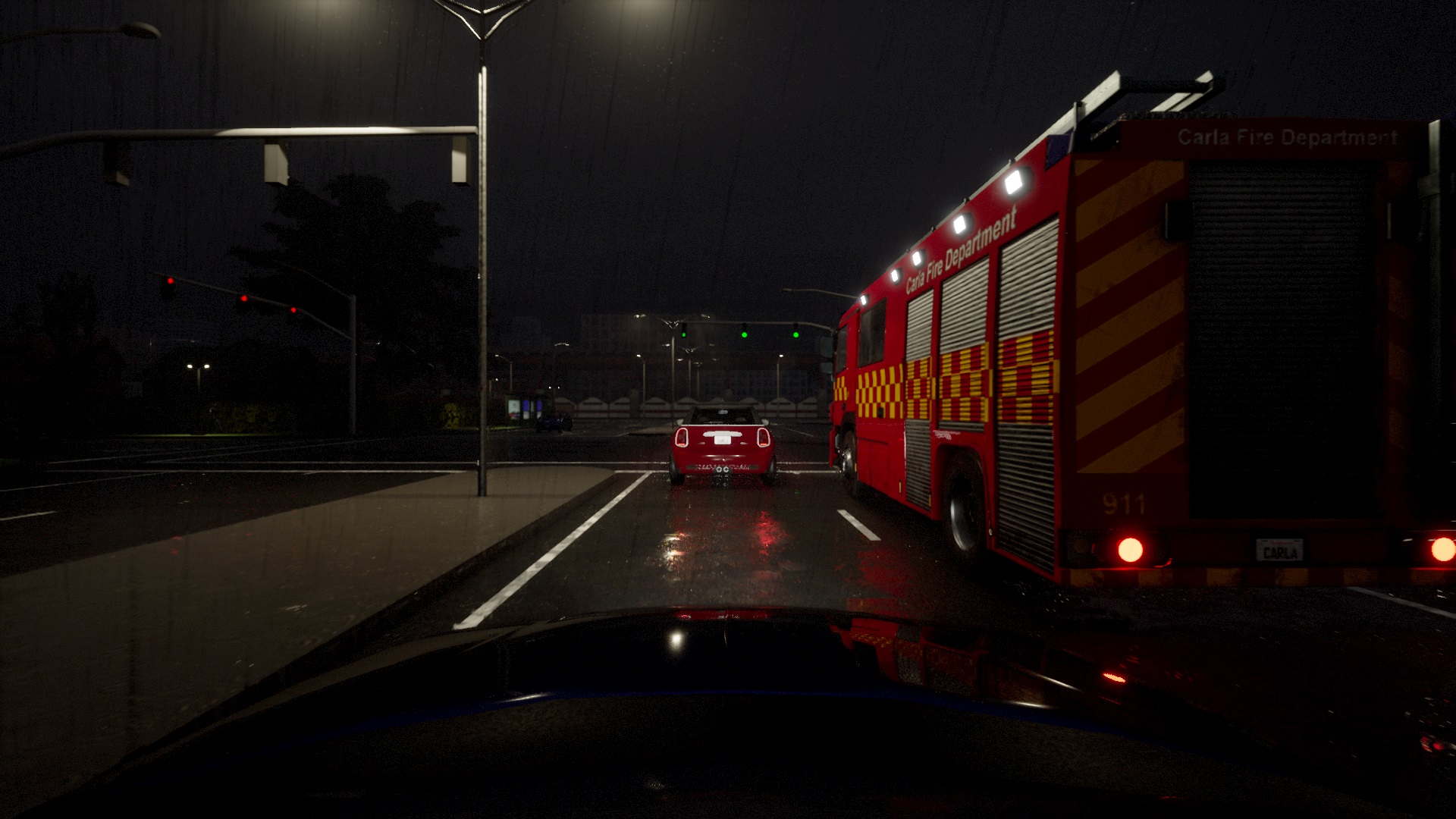} &
        \includegraphics[width=.325\textwidth]{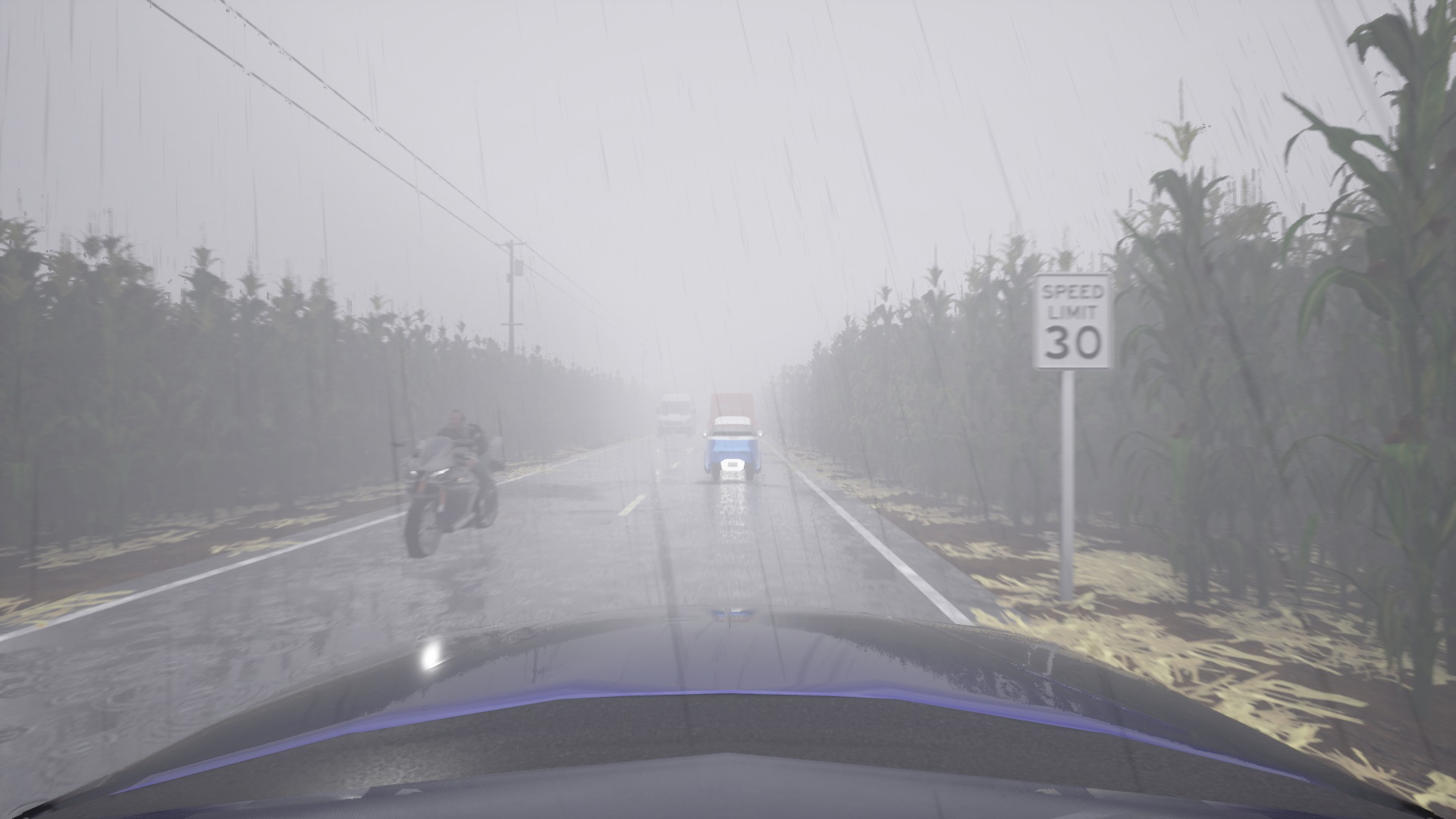} &
        \includegraphics[width=.325\textwidth]{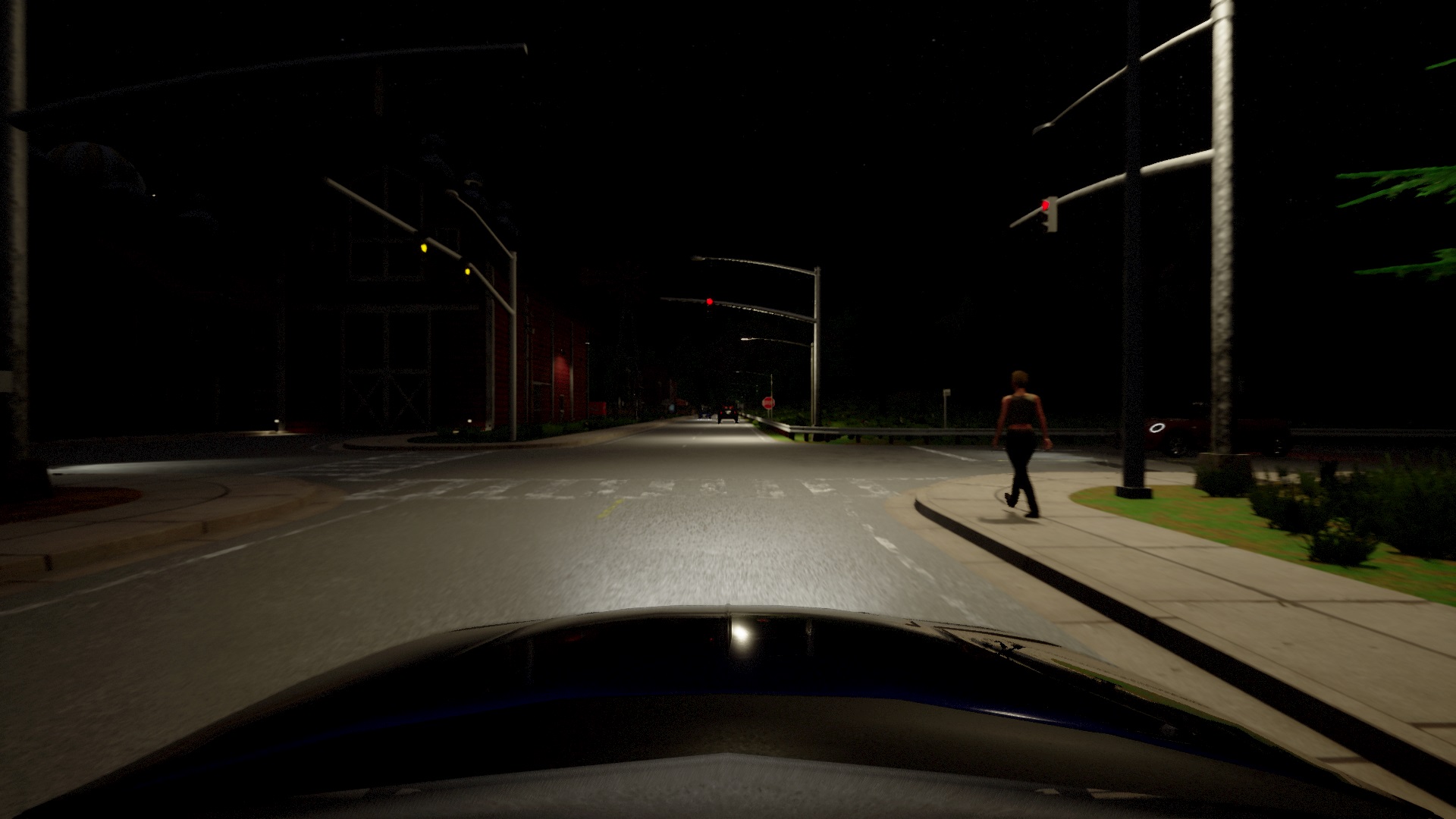} \\  
        (a) & (b) & (c) \\
        \includegraphics[width=.325\textwidth]{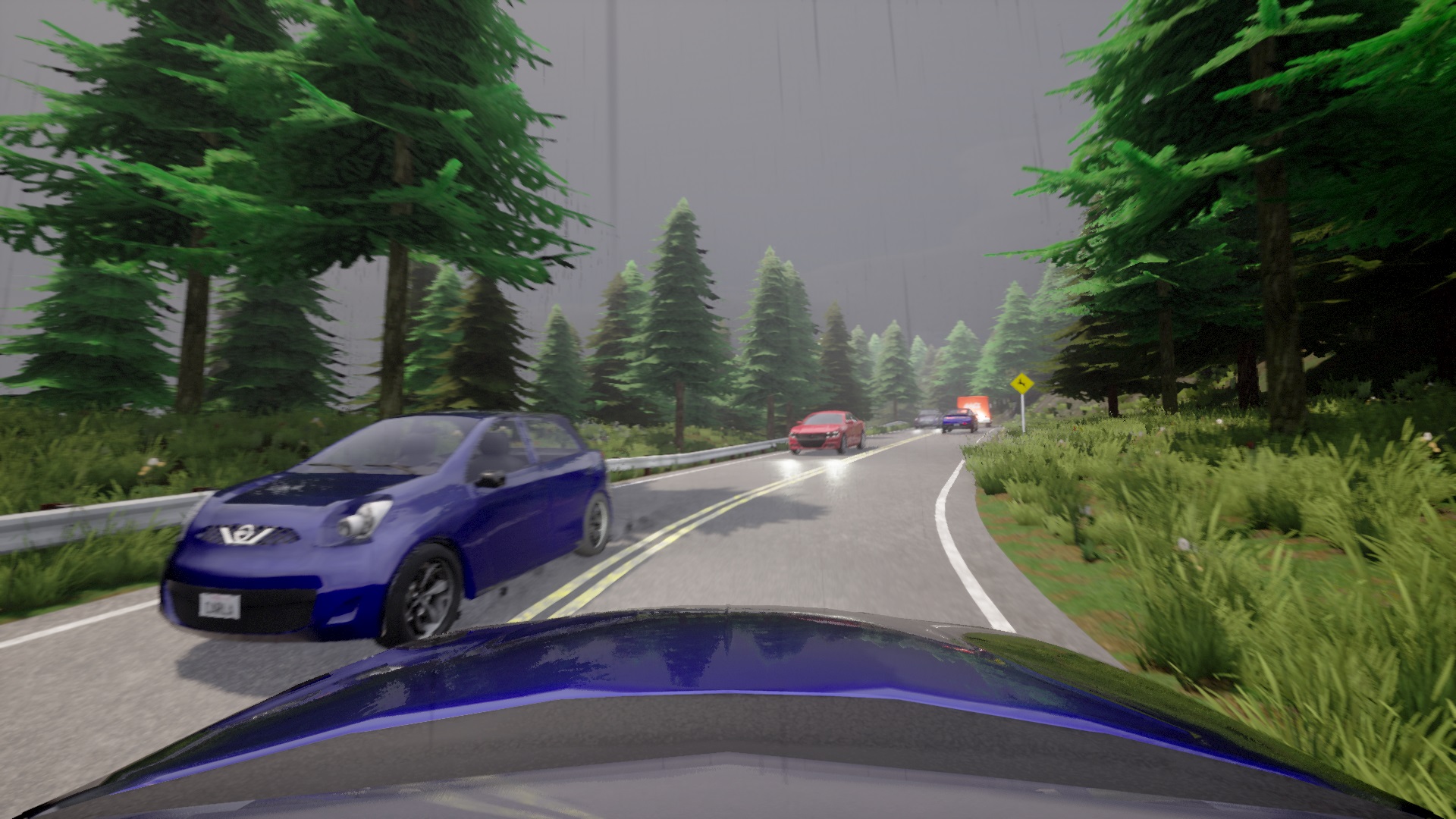} & 
        \includegraphics[width=.325\textwidth]{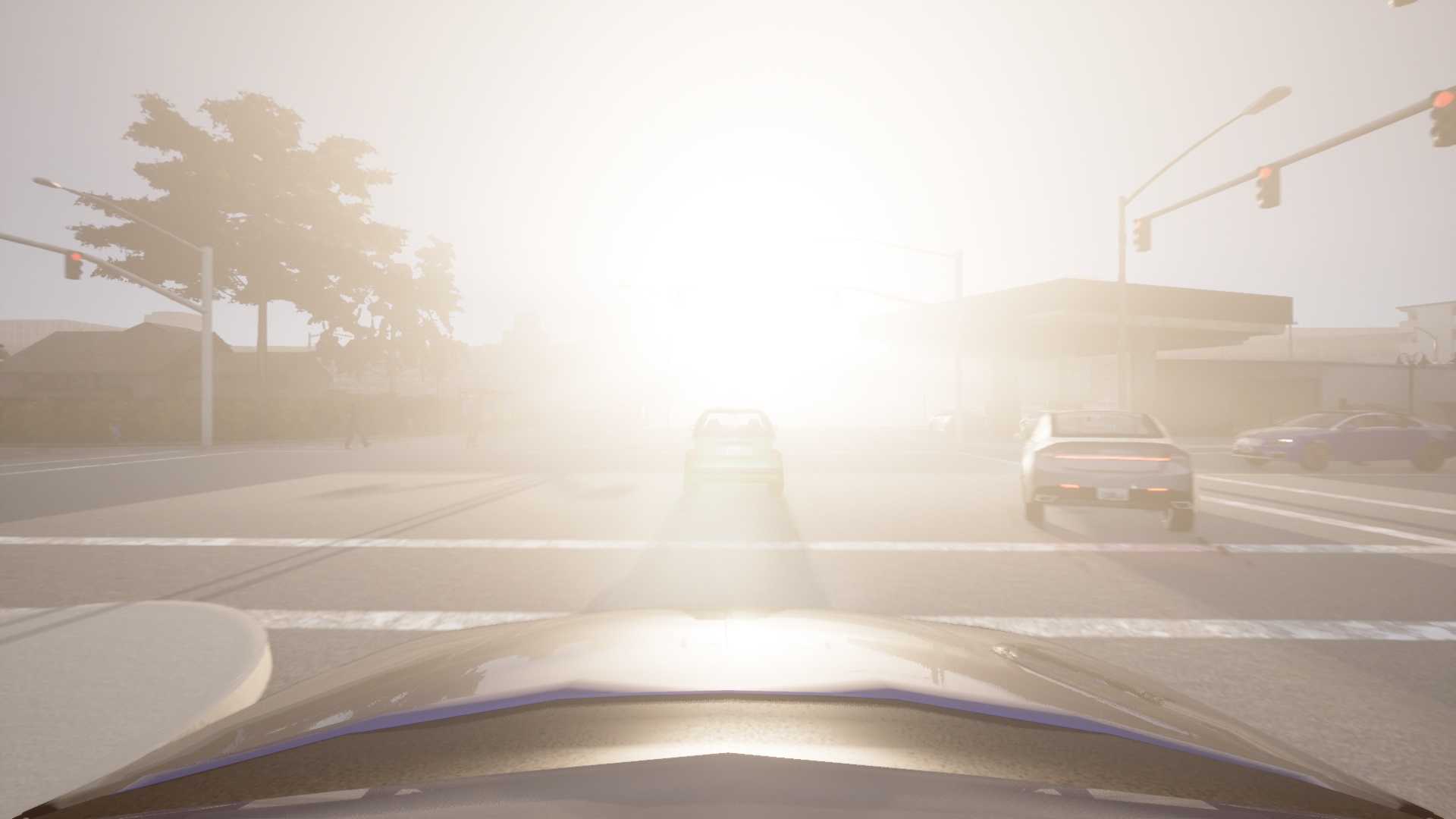} &
        \includegraphics[width=.325\textwidth]{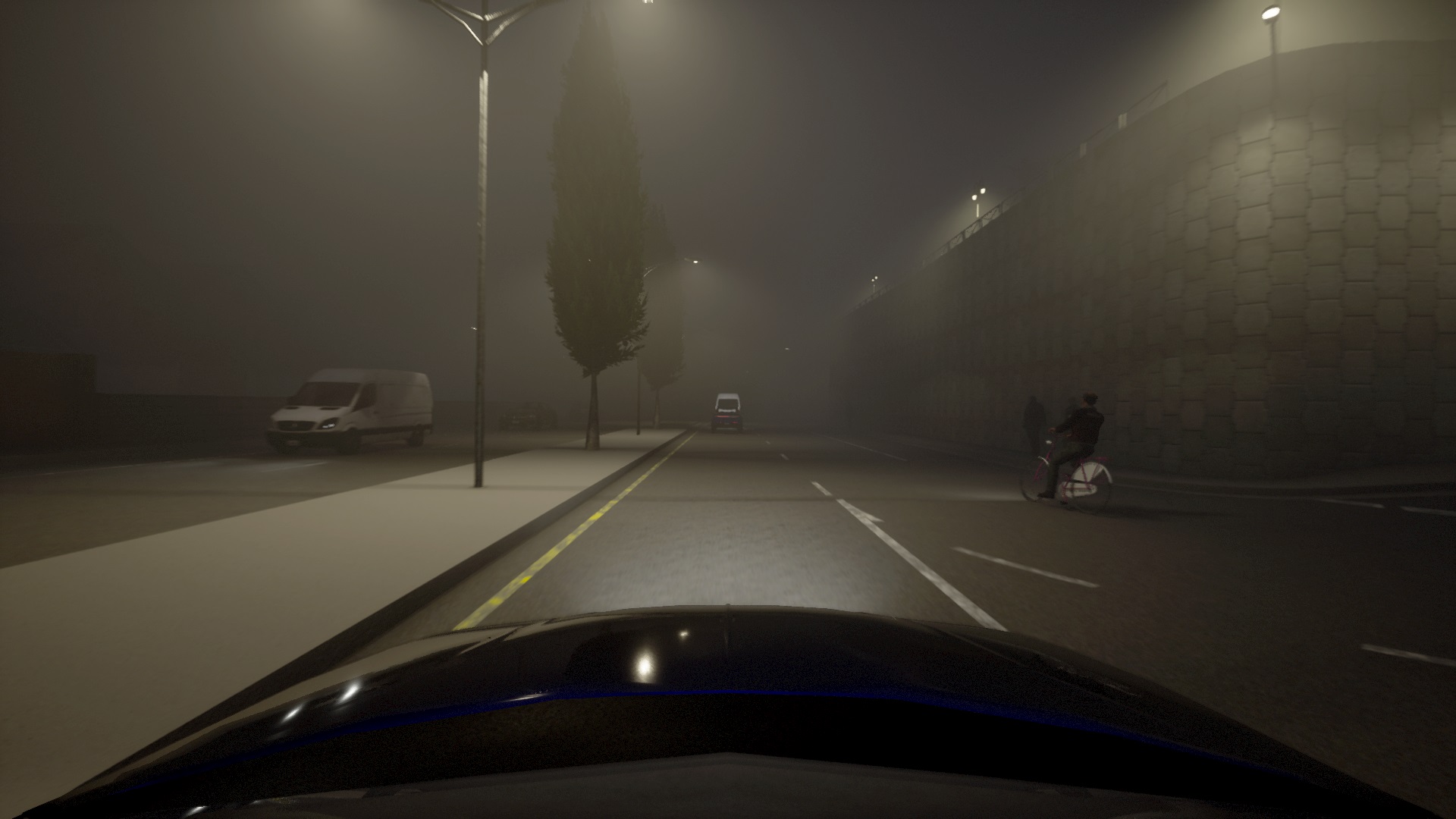} \\ 
        (d) & (e) & (f)
    \end{tabular}
    \captionof{figure}{Camera samples from Adver-City showcasing its weather diversity. (a) Heavy rain night at an urban intersection. (b) Foggy heavy rain day at a rural straight non-junction. (c) Clear night at a rural intersection. (d) Soft rain day at a rural curved non-junction. (e) Glare day at an urban intersection. (f) Foggy night at an urban non-junction.}
    \label{fig:scenarios}
\endgroup
\end{figure*}

To address this gap, we introduce Adver-City, the first open-source CP dataset focused on adverse weather, created using the CARLA simulator~\cite{dosovitskiy_2017_carla} and OpenCDA framework~\cite{xu_2021_opencda}. The dataset includes $110$ diverse scenarios across six distinct weather conditions: clear weather, glare, soft rain, heavy rain, fog and foggy heavy rain (Figure~\ref{fig:scenarios}). It is the first synthetic CP dataset to feature glare and incorporate real-life data on adverse weather and poor visibility conditions in its scenario design. Each scene includes three CAVs and two RSUs, and multi-modal data from LiDARs, RGB and semantic cameras, GNSS, and IMUs, supporting tasks like object detection, tracking, and semantic segmentation.

Our key contributions could then be summarized as:
\begin{itemize}
    \item Adver-City, the first open-source synthetic CP dataset focused on adverse weather conditions;
    \item Glare scenarios, which have not yet been explored by CP datasets;
    \item A scenario design approach based on real-life data that accounts for challenging conditions involving adverse weather and poor visibility;
    \item A sensor suite balancing performance and real-world feasibility, supported by sensor configurations commonly used in practice and enabling Sim2Real approaches.
\end{itemize}

This paper is organized as follows: Section \ref{sec:related_work} presents and discusses synthetic multi-modal CP datasets, Section \ref{sec:dataset} describes the Adver-City dataset in regard to its scenarios, generation and statistics, Section \ref{sec:benchmark} presents the results of benchmarks performed with the dataset, Section \ref{sec:limitations} discusses limitations regarding the realism of the dataset, and Section \ref{sec:conclusions} discusses possible directions to be taken in the future.
\section{Related Work}
\label{sec:related_work}

Collaborative Perception has recently gained considerable attention, leading to the development of various datasets. Some, like DAIR-V2X~\cite{yu_2022_dairv2x}, V2V4Real~\cite{xu_2023_v2v4real} and TUMTraf~\cite{zimmer_2024_tumtraf}, are based on real-world data, while others use synthetic data from simulators such as CARLA~\cite{dosovitskiy_2017_carla} and LidarSIM~\cite{manivasaga_2020_lidarsim}. Among the synthetic datasets, some focus solely on LiDAR data, such as V2V-Sim~\cite{wang_2020_v2vnet} and CODD~\cite{arnold_2022_fast}, while others provide multi-modal data from both LiDARs and cameras, as shown in Table~\ref{tab:datasets} and discussed here.

\begin{table*}%[!hbt]
\centering
\resizebox{0.99\textwidth}{!}{\begin{tabular}{c|cccccc|cccc|cc}
\toprule
\textbf{Dataset} & \textbf{V2X} & \textbf{\begin{tabular}[c]{@{}c@{}}View-\\ points\end{tabular}} & \textbf{Scenes} & \textbf{\begin{tabular}[c]{@{}c@{}}Frames\\ (k)\end{tabular}} & \textbf{\begin{tabular}[c]{@{}c@{}}Annota-\\tions (k)\end{tabular}} & \textbf{\begin{tabular}[c]{@{}c@{}}Cate-\\ gories\end{tabular}} & \textbf{RGB} & \textbf{Lidar} & \textbf{Semantic} & \textbf{\begin{tabular}[c]{@{}c@{}}GNSS\\\& IMU\end{tabular}} & \textbf{Weather} & \textbf{\begin{tabular}[c]{@{}c@{}}Open\\ Source\end{tabular}} \\
\midrule
OPV2V \cite{xu_2021_opv2v} & V2V    & 2-7& 73   & 11     & 233 & 1     & $\checkmark$    & $\checkmark$ & $\times$     & $\checkmark$& $\times$    & $\checkmark$     \\
V2X-Sim \cite{li_2022_v2xsim}    & V2X    & 2-7& 100  & 10     & -   & 23    & $\checkmark$    & $\checkmark$ & $\checkmark$    & $\checkmark$& $\times$    & $\times$ \\
DOLPHINS \cite{mao_2022_dolphins}   & V2X    & 3  & 6    & 42     & 293 & 2     & $\checkmark$    & $\checkmark$ & $\times$     & $\times$ & $\times$    & $\checkmark$     \\
DeepAccident \cite{wang_2023_deepaccident}    & V2X    & 5  & 691  & 57     & -   & 6     & $\checkmark$    & $\checkmark$ & $\checkmark$    & $\checkmark$& $\checkmark$   & $\times$\\
SCOPE \cite{gamerdinger_2024_scope} & V2X    & 3-24     & 44   & 17     & 575 & 5     & $\checkmark$    & $\checkmark$ & $\checkmark$    & $\times$ & $\checkmark$   & $\times$\\
\midrule
Adver-City & V2X    & 5  & 110  & 24     & 889 & 6     & $\checkmark$    & $\checkmark$ & $\checkmark$    & $\checkmark$& $\checkmark$   & $\checkmark$    \\
\bottomrule
\end{tabular}}
\caption{Comparison between multi-modal CP datasets generated using CARLA~\cite{dosovitskiy_2017_carla}. The number of annotations for V2X-Sim and DeepAccident is not provided by their authors. All datasets are publicly available, yet only the Open Source datasets provide scenario generation code.}
\label{tab:datasets}
\end{table*}

OPV2V~\cite{xu_2021_opv2v} is arguably the most used CP dataset to date, serving as a benchmark for several collaborative models, including Coalign~\cite{lu_2023_robust}, Where2Comm~\cite{hu_2022_wherecomm} and CoBEVT~\cite{xu_2022_cobevt}. Built with the open-source OpenCDA framework~\cite{xu_2021_opencda}, it has inspired variations such as OPV2V-H~\cite{lu_2024_heal}, OPV2V+~\cite{hu_2023_collaboration} and OPV2VH+~\cite{yue_2024_communication}. However, it does not contain RSUs, includes only one object category (cars), and features only clear weather scenarios.

Another widely used dataset is V2X-Sim \cite{li_2022_v2xsim}, which was used in models such as STAR~\cite{li_2022_multirobot}, SyncNet~\cite{lei_2022_latency} and DiscoNet~\cite{li_2021_learning}. It follows the annotation schema of the single-vehicle NuScenes dataset~\cite{caesar_2020_nuscenes}, featuring RSUs, semantic segmentation ground truths, and multiple object categories. However, it lacks adverse weather scenes and does not provide the source code for scenario generation, limiting its modifiability. 

The DOLPHINS~\cite{mao_2022_dolphins} dataset, while open-source, is less commonly used and built on a less robust codebase than OpenCDA's~\cite{xu_2021_opencda}. In addition, it includes only six clear weather scenarios with data captured from a LiDAR sensor and a single front-facing camera per agent.

Focusing on accident prediction, DeepAccident \cite{wang_2023_deepaccident} features a large number of scenarios with diverse object categories, including pedestrians and two-wheeled vehicles. The scenes, selected based on real-world crash reports, include a few environmental conditions -- $37\%$ are rainy and $16.3\%$ occur at night. However, the scenario generation code is not publicly available and, due to its focus on accident prediction, the scenes are short, averaging less than $9$ seconds in length.

The recent SCOPE dataset \cite{gamerdinger_2024_scope} performs image and point cloud augmentation on top of CARLA's sensor data to provide more realistic weather. It includes data from digital twin maps of German cities and a solid-state LiDAR among its sensors. Despite its focus on adverse weather, only $44.2\%$ of its scenes feature fog or rain, and only $25.8\%$ take place at night. Additionally, neither the scenario generation code nor the weather augmentation algorithms have been made publicly available.
\section{Adver-City Dataset}
\label{sec:dataset}

\subsection{Scenarios}
\label{subsec:scenarios}

Inspired by DeepAccident~\cite{wang_2023_deepaccident}, we selected scenarios based on comprehensive real-world crash data from the US National Highway Traffic Safety Administration (NHTSA). Even though accidents involving AVs are fundamentally different from those involving only non-autonomous vehicles (over $90\%$ of crashes are caused by driver error~\cite{nhtsa_2018_critical}), data on AV crashes remain scarce. Public AV crash reports document only $30$~\cite{liu_2024_analysis} or fewer~\cite{liu_2021_crash} AV crashes, compared to the latest NHTSA report~\cite{nhtsa_2019_statistics}, which records over $5$ million accidents for non-autonomous vehicles.

An NHTSA report from 2007~\cite{nhtsa_2007_typology} defined a typology consisting of $37$ pre-crash scenarios based on vehicle movements and the critical event occurring immediately before a crash, presenting detailed statistics on each scenario. Then, in 2019, another NHTSA report~\cite{nhtsa_2019_statistics} updated the typology to a total of $36$ pre-crash scenarios, clustering them in $9$ scenario groups (Table \ref{tab:groups}) and providing up-to-date statistics on each group.

\begin{table}
\centering
\resizebox{0.478\textwidth}{!}{\begin{tabular}{l||c|c||c|c}
\toprule
\textbf{Scenario Group} & \textbf{Weather*}  & \textbf{\begin{tabular}[c]{@{}c@{}}Bad\\Visibility*\end{tabular}} & \textbf{\begin{tabular}[c]{@{}c@{}}Fatality\\ Rate (\%)\end{tabular}} & \textbf{\begin{tabular}[c]{@{}c@{}}Vehicles\\ per Crash\textsuperscript{\textdagger}\end{tabular}} \\ \midrule
Animal  & 10 & 1 & 0.03 & 1.02  \\
Rear-End& 12 & 1 & 0.07 & \underline{\textbf{2.19}}  \\
Lane Change & 10 & 2 & 0.12 & 2.00  \\
Crossing Paths & 11 & \underline{\textbf{7}}     & 0.35 & \underline{\textbf{2.04}}  \\
Control Loss& \underline{\textbf{44}} & 1 & 0.95 & 1.16  \\
Pedalcyclist& 5  & \underline{\textbf{7}}     & 1.08 & 1.05  \\
Road Departure & 13 & 2 & 1.19 & 1.01  \\
Opposite Direction     & \underline{\textbf{15}} & 4 & \underline{\textbf{3.23}} & 1.88  \\
Pedestrian  & 13 & \underline{\textbf{10}}    & \underline{\textbf{5.29}} & 1.07 \\
\bottomrule
\end{tabular}}
\caption{Statistics from the latest NHTSA report~\cite{nhtsa_2019_statistics} on pre-crash scenario groups. The highest two values in each column are highlighted for convenience. *Percentage of crashes in each scenario group that had adverse conditions as the main factor. \textsuperscript{\textdagger}Average number of vehicles per crash in each group.}
\label{tab:groups}
\end{table}

Since the main problems for object detection under adverse weather conditions may be attributed to poor visibility and sensor performance degradation \cite{rothmeier_2023_badday}, we filtered the pre-crash scenarios groups by focusing on those that have, among their main causes, errors in perception (i.e. identification of an object, such as a car or pedestrian) instead of errors in control (often caused by driver impairment, driver inexperience or speeding). Moreover, we focused on scenario groups with the highest fatality rates, strongly influenced by adverse weather conditions and poor visibility.

Even though $44\%$ of the crashes in the `Control Loss' group had adverse weather as the main factor, most of them had errors in control as their main cause. This may be attested by the fact that the average number of vehicles per crash is close to $1$ and that, by definition, there are no external entities (e.g., animals, cyclists or pedestrians) involved in crashes in this group. A similar consideration may be made for the `Road Departure' group. 

Therefore, the most relevant groups, based on our criteria, are `Pedestrian', `Pedalcyclist', `Opposite Direction' and `Crossing Paths'. The last two gain even more relevance given that they have the highest fatality rate for motorcyclists, according to a 2020 NHTSA report focused exclusively on crashes involving motorcycles~\cite{nhtsa_2020_motorcycle}.

From these four groups, we looked back at the detailed scenario statistics of the 2007 report~\cite{nhtsa_2007_typology} and collected the following data regarding the road configurations commonly present in these groups:

\begin{itemize}
    \item \textbf{Pedestrian}: crashes happened mostly on urban roads ($65\%$), with crashes where the vehicle maneuvered happening mostly at intersections ($81\%$) and without maneuver at non-junctions ($55\%$);
    \item \textbf{Pedalcyclist}: evenly split between urban ($54\%$) and rural ($46\%$) roads, but happening more commonly at intersections ($68\%$);
    \item \textbf{Opposite Direction}: mostly happening at rural roads ($65\%$) and non-junctions ($87\%$), with a significant percentage happening at non-level ($34\%$) or curved roads ($39\%$);
    \item \textbf{Crossing Paths}: even split between rural ($53\%$) and urban ($47\%$) roads, mostly happening at intersections ($63\%$).
\end{itemize}

In light of this data, we selected five road configurations that cover all of the above cases and modeled them on the following CARLA maps~\cite{dosovitskiy_2017_carla}:

\begin{enumerate}
    \item \textbf{Urban intersection} (straight and level) [Town03];
    \item \textbf{Urban non-junction} (straight and level) [Town03];
    \item \textbf{Rural intersection} (straight and level) [Town07];
    \item \textbf{Rural straight non-junction} (level) [Town07];
    \item \textbf{Rural curved non-junction} (non-level) [Town07].
\end{enumerate}

We then drew inspiration from the single-vehicle Oxford RobotCar Dataset~\cite{maddern_2017_oxford}, wherein a specific route was traversed twice a week for a year, capturing diverse scenes along the same path. Building on this, we introduced variations to the road configurations to simulate how these locations change across days. The first variation was object density, with two levels: \textbf{sparse}, featuring a fixed number of vehicles and pedestrians per road configuration, and \textbf{dense}, which increases the number of pedestrians by $2.33\times$ and vehicles by $2.67\times$, amplifying occlusion. The second variation was weather, with eleven weather/daytime conditions simulated in CARLA:

\begin{itemize}
    \item \textbf{Clear weather} (Day and Night);
    \item \textbf{Soft rain} (Day and Night);
    \item \textbf{Heavy rain} (Day and Night);
    \item \textbf{Fog} (Day and Night);
    \item \textbf{Foggy Heavy ran} (Day and Night);
    \item \textbf{Glare} (Day).
\end{itemize}

By combining $11$ weather and $2$ density levels across $5$ road configurations, we created a total of $110$ unique scenarios. This setup enables detailed comparisons between environmental factors, such as the impact of rain intensity across different locations or the effects of increased object density at intersections. Despite stemming from only five road configurations, each scenario was simulated independently so that, for a given road configuration and density pairing, the agents' behavior varies slightly, resulting in a diverse set of simulations.

\subsection{Dataset Generation}
\label{subsec:generation}

The Adver-City dataset was generated in the CARLA simulator~\cite{dosovitskiy_2017_carla} with the OpenCDA~\cite{xu_2021_opencda} framework. Each scenario has data from five distinct viewpoints, as illustrated in Figure~\ref{fig:povs}: two RSUs, the ego vehicle, and two CAVs. Each viewpoint is equipped with the sensors listed in Table~\ref{tab:sensors}. 

\begingroup
\renewcommand{\arraystretch}{0}
\begin{figure}%[!hbt]
    \hspace*{-0.25cm}
    \begin{tabular}{c@{}c}
   \includegraphics[width=.237\textwidth]{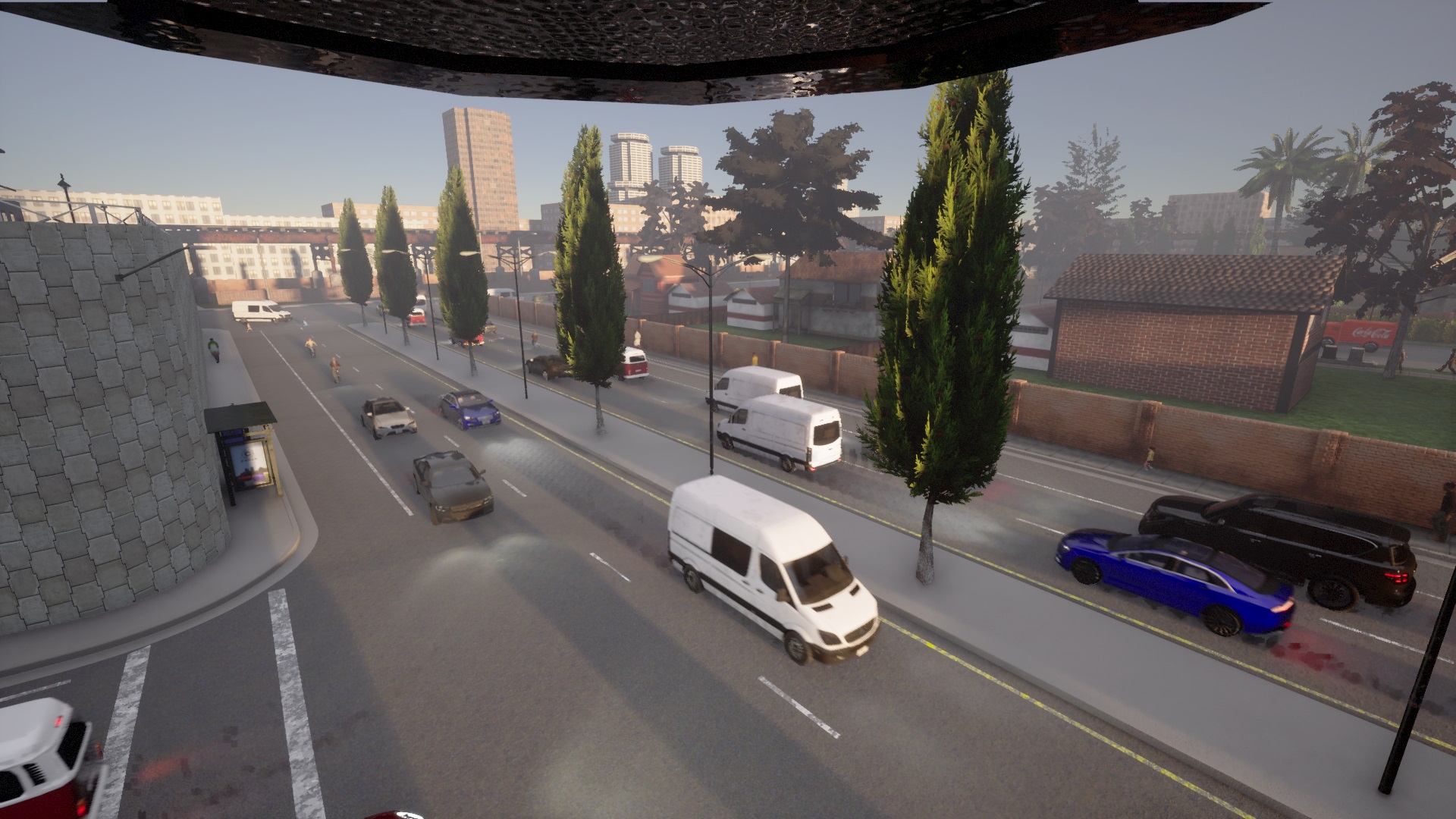} & 
   \includegraphics[width=.237\textwidth]{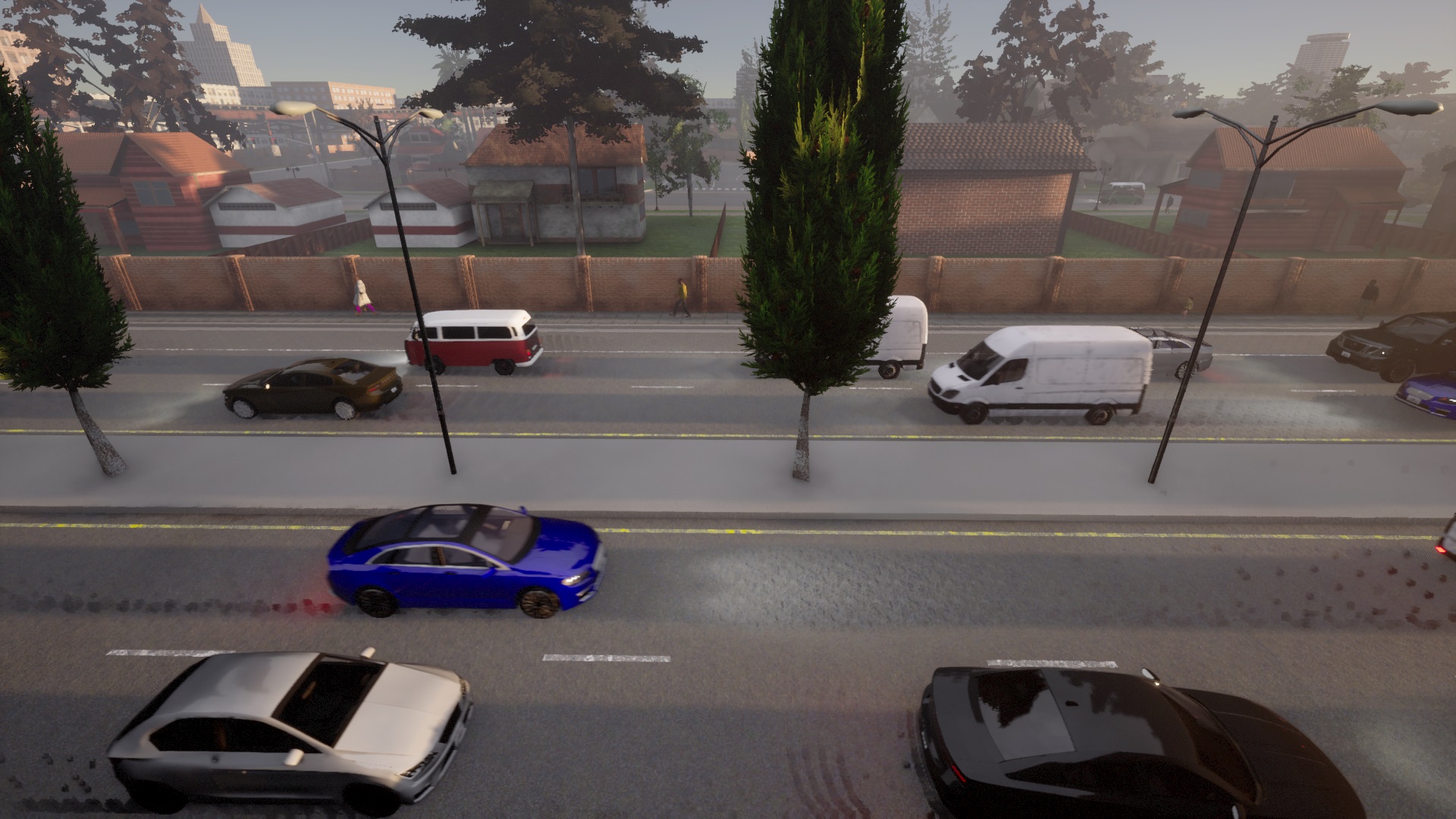} \\
   \multicolumn{2}{c}{
  \includegraphics[width=.474\textwidth]{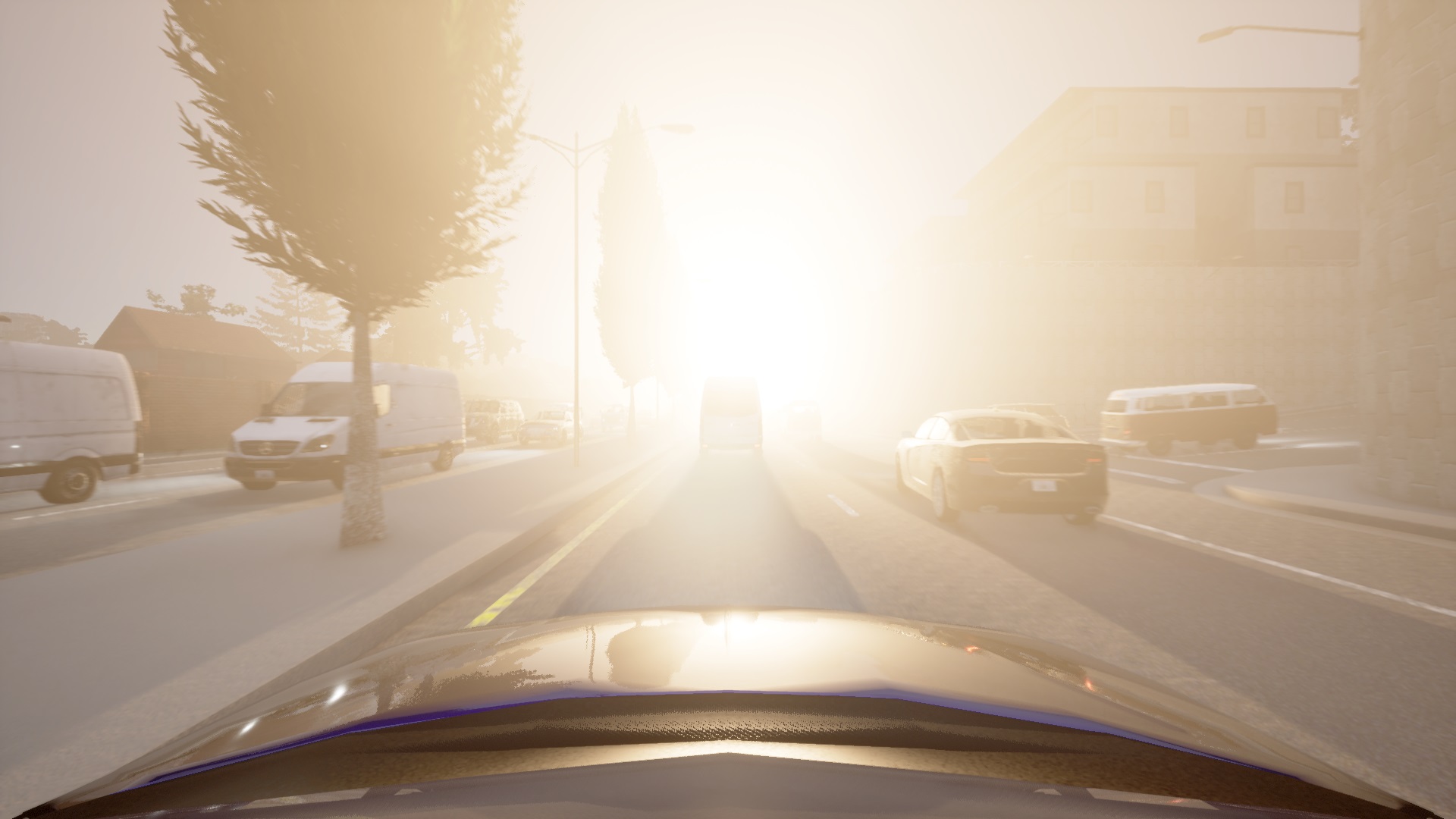}
   }\\
   \includegraphics[width=.237\textwidth]{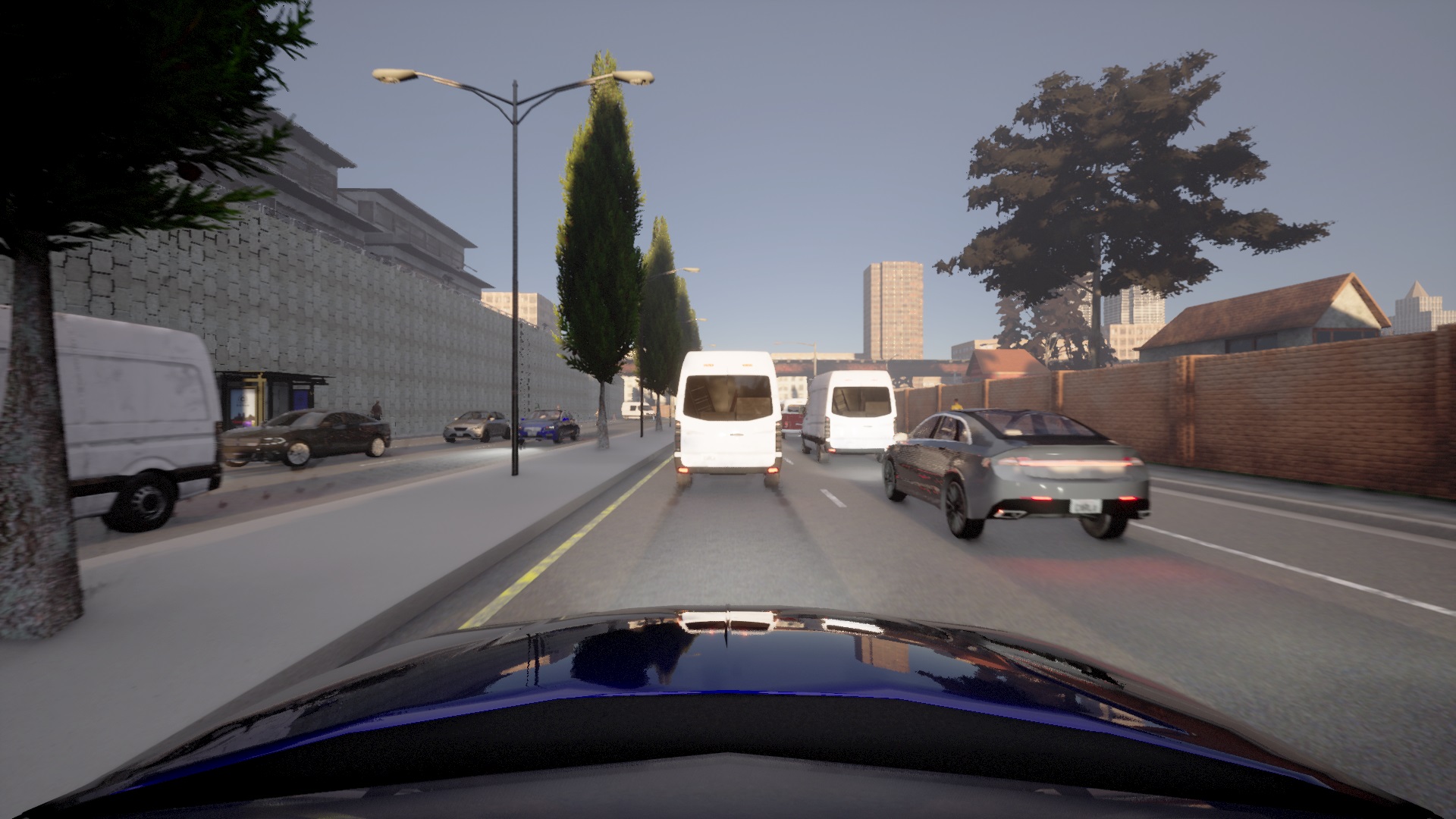} &
   \includegraphics[width=.237\textwidth]{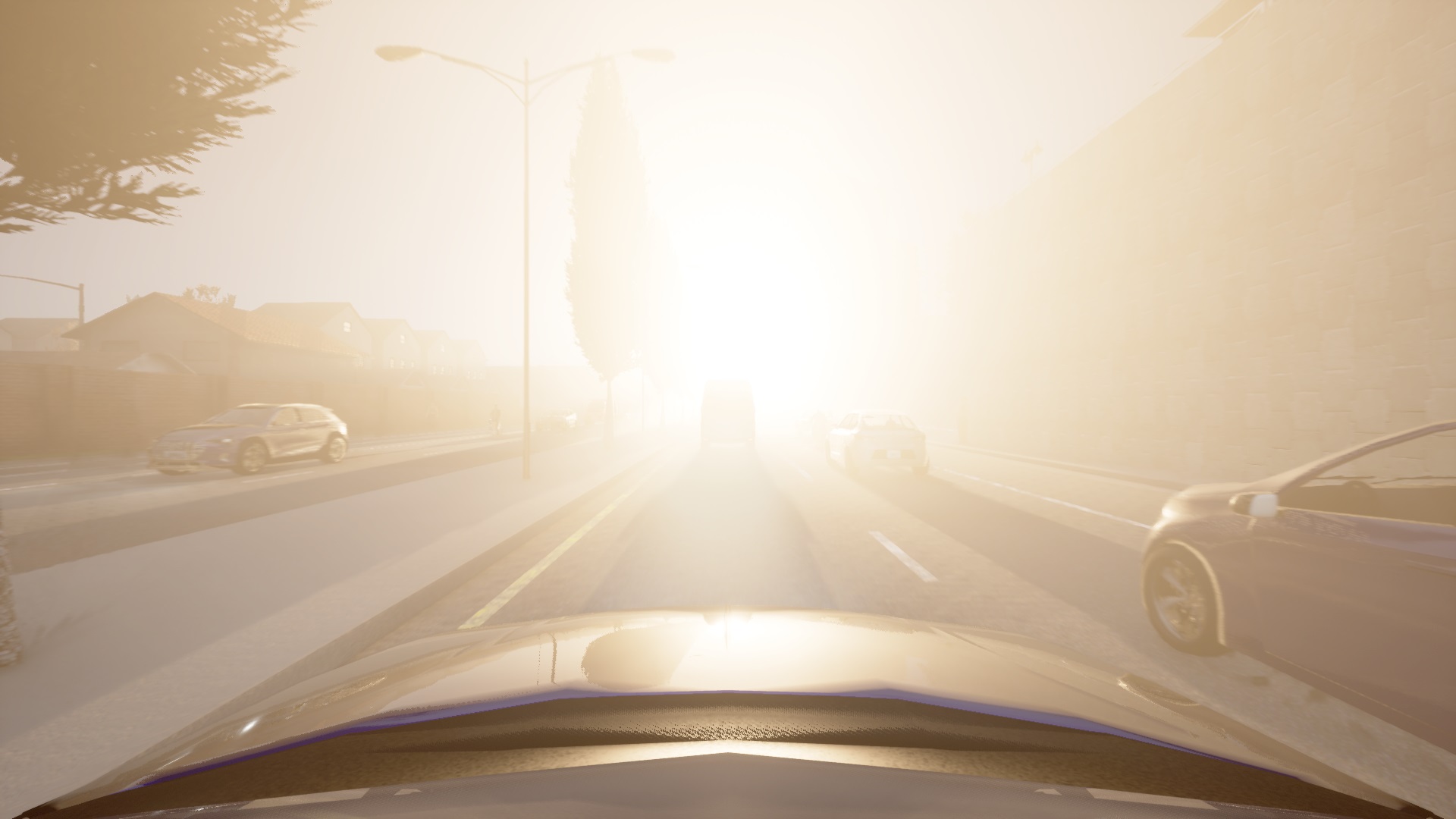} 
    \end{tabular}
    \caption{Camera samples from the different viewpoints used in the Adver-City dataset. Scenario: glare day at urban non-junction. Top: RSUs, center: ego AV, bottom: connected AVs. Despite the ego's front camera being heavily affected by glare, the RSUs remain unaffected, providing valuable information about the ego's surroundings.}
    \label{fig:povs}
\end{figure}
\endgroup

\begin{table}%[!hbt]
\centering
\resizebox{0.478\textwidth}{!}{\begin{tabular}{ll}
\toprule
\textbf{Sensors}    & \textbf{Details} \\ 
\midrule
$4$x RGB Camera  & $1920\times1080$, $100^\circ$ horizontal FOV \\
\begin{tabular}[c]{@{}l@{}}$4$x Semantic\\Camera\end{tabular}   & $1920\times1080$, $100^\circ$ horizontal FOV \\
$1$x LiDAR   & \begin{tabular}[c]{@{}l@{}}$32$ channels, $1.2$M points/sec, $10$Hz capture\\ frequency, $200$m range, $-25^\circ$ to $15^\circ$ vertical FOV*\end{tabular} \\
GNSS \& IMU& \begin{tabular}[c]{@{}l@{}}$3$e$-6$ std. dev. for latitude and longitude's noise\\models, $0.05$ std. dev. for altitude's noise model\\ \end{tabular} \\ 
\bottomrule
\end{tabular}}
\caption{Sensor specifications for the AVs used in Adver-City scenarios. The RSUs are not equipped with GNSS \& IMU but otherwise follow the same specification. *RSUs positioned above ground have a $-25^\circ$ to $-1^\circ$ vertical FOV.}
\label{tab:sensors}
\end{table}

The sensor configuration was selected to balance detection performance with real-world feasibility. We used $1080$p ($1920\times1080$) cameras and $32$-channel LiDARs while omitting depth cameras. The $1080$p cameras offer more detailed imagery compared to lower resolution options and are relatively affordable and widely available. The $32$-channel LiDARs offer less detail than $64$-channel models but are also more accessible. Although depth cameras may provide unique information, they are rarely used in practice. None of the real-world datasets discussed in Section~\ref{sec:related_work} use depth cameras; instead, they use LiDARs with fewer than $64$ channels and cameras with a resolution of $1080$p or higher. 

\begin{figure}%[!hbt]
    \includegraphics[width=.474\textwidth]{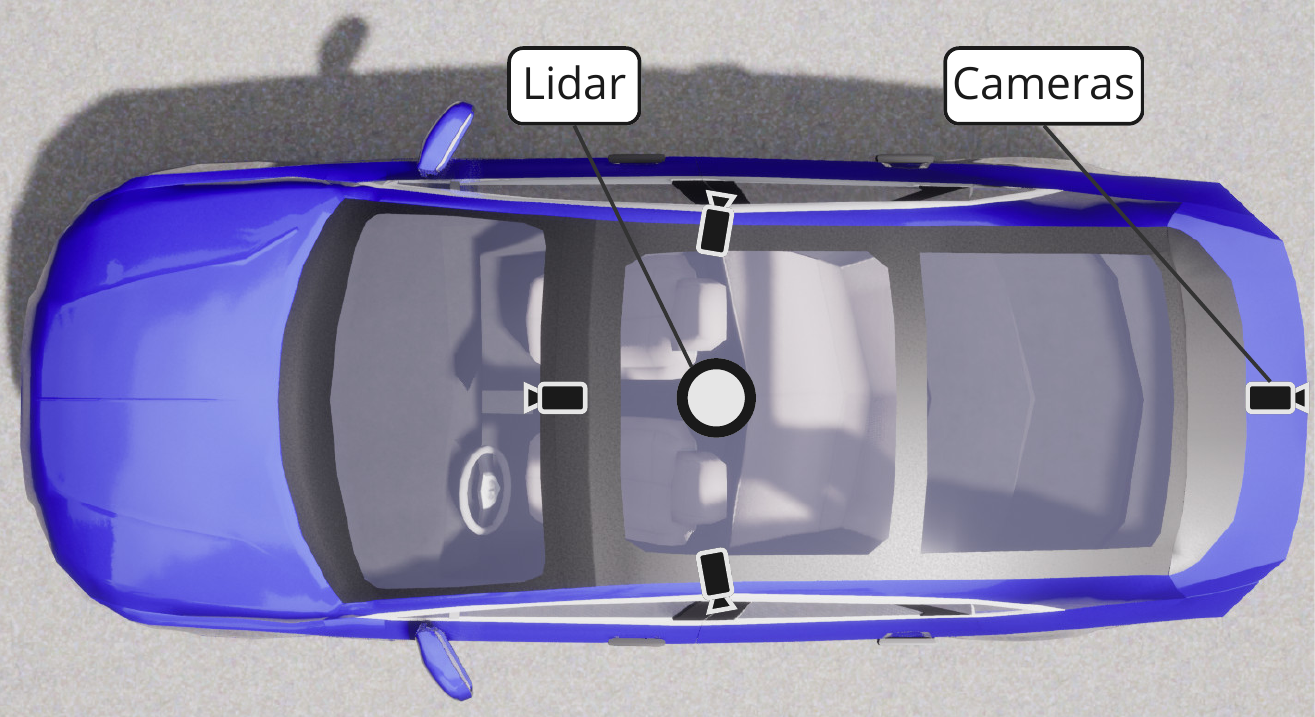}
    \caption{Sensor positioning in CAVs. Camera positioning indicates both RGB and semantic segmentation cameras. Side-facing cameras are tilted backwards by $10^\circ$ to reduce motion blur effects.}
    \label{fig:sensors}
\end{figure}

The AVs' sensors are positioned as illustrated in Figure~\ref{fig:sensors}: the frontal camera is mounted on the windshield, the rear camera is located at the back of the trunk, side-facing cameras are placed at the top of the B-pillars (tilted backwards by $10^\circ$ to reduce motion blur), and the LiDAR is positioned at the top of the vehicle. RSUs are either attached to tall structures (such as walls, towers, traffic light poles, or light posts) or positioned in high-visibility spots along the road. In non-junction scenarios, one CAV is positioned ahead of the ego vehicle, travelling in the same direction, while the other CAV is also ahead but travelling in the opposite direction. In intersection scenarios, CAVs approach the intersection from different roads compared to the ego vehicle. More details on the placement of RSUs and CAVs for each road configuration are provided in the Supplementary Material.

Data was captured at $10$Hz and adheres to an annotation schema compatible with OPV2V \cite{xu_2021_opv2v}, with the main difference relating to the categories in our dataset. Unlike OPV2V, which only included cars and did not require category annotations for bounding boxes, our dataset has six categories: cars, pedestrians, vans, trucks, bicycles, and motorcycles. 

\subsection{Dataset Statistics}

\begin{figure*}%[!hbt]
\centering
\begin{subfigure}{.245\textwidth}
    \centering
    \includegraphics[width=\textwidth]{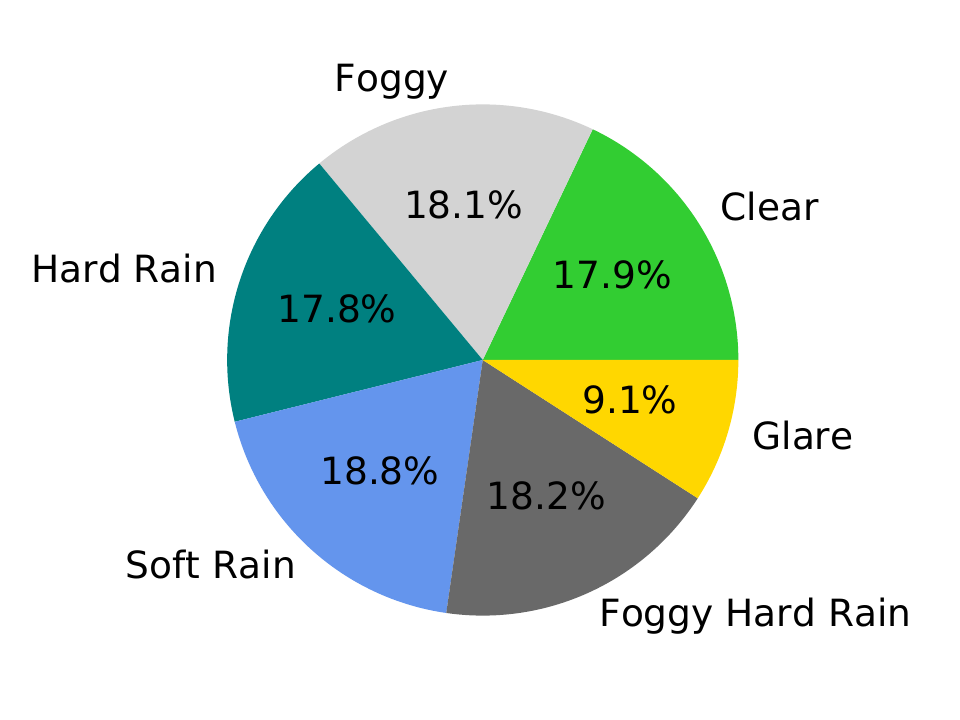}
    \caption{}
    \label{subfig:weather}
\end{subfigure}
\begin{subfigure}{.245\textwidth}
    \centering
    \includegraphics[width=\textwidth]{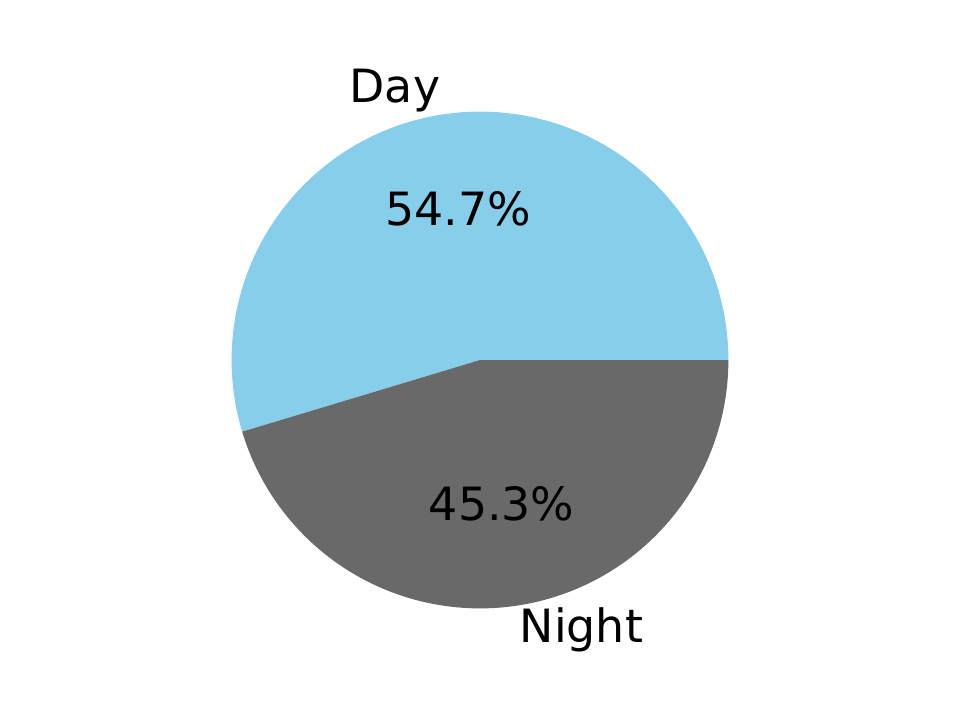}
    \caption{}
    \label{subfig:time_of_day}
\end{subfigure}
\begin{subfigure}{.245\textwidth}
    \centering
    \includegraphics[width=\textwidth]{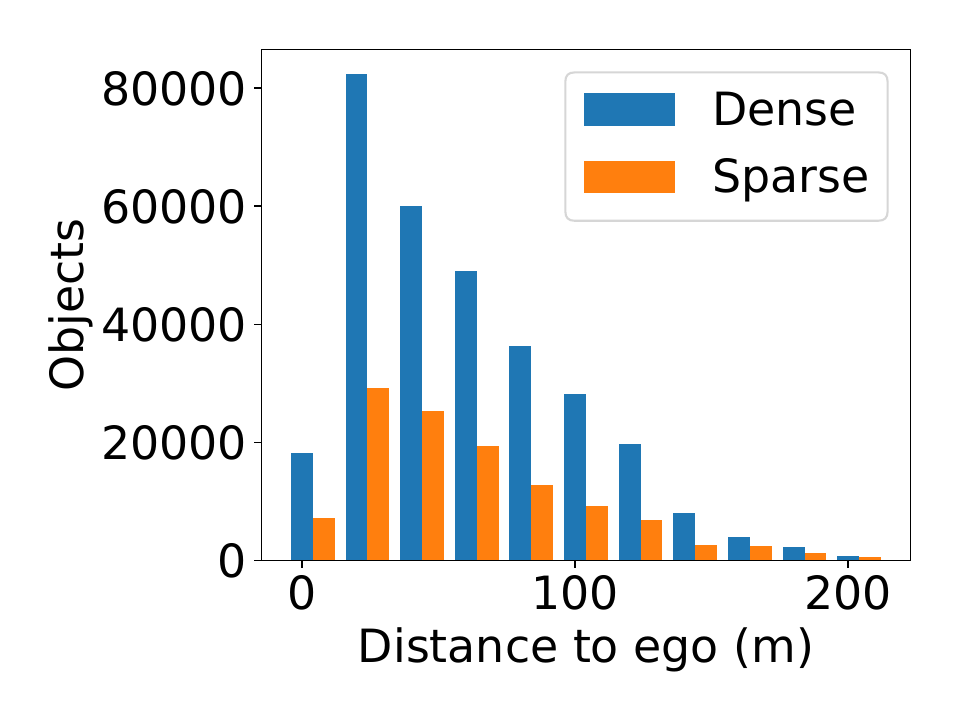}
    \caption{}
    \label{subfig:density}
\end{subfigure}
\begin{subfigure}{.245\textwidth}
    \centering
    \includegraphics[width=\textwidth]{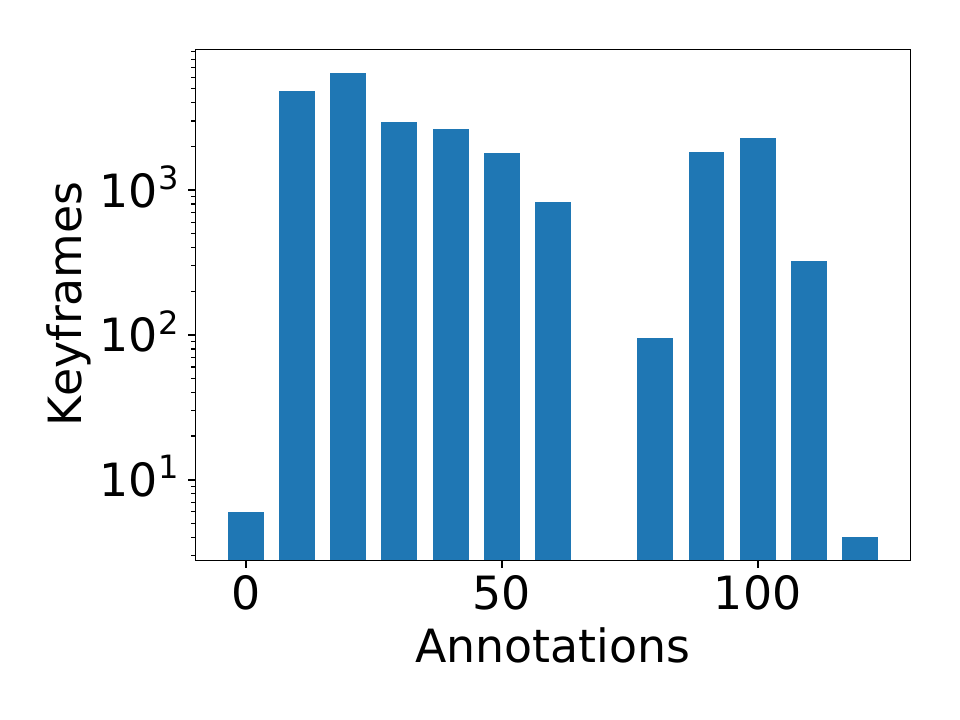}
    \caption{}
    \label{subfig:annotations}
\end{subfigure}
\begin{subfigure}{.38\textwidth}
    \centering
    \includegraphics[width=\textwidth]{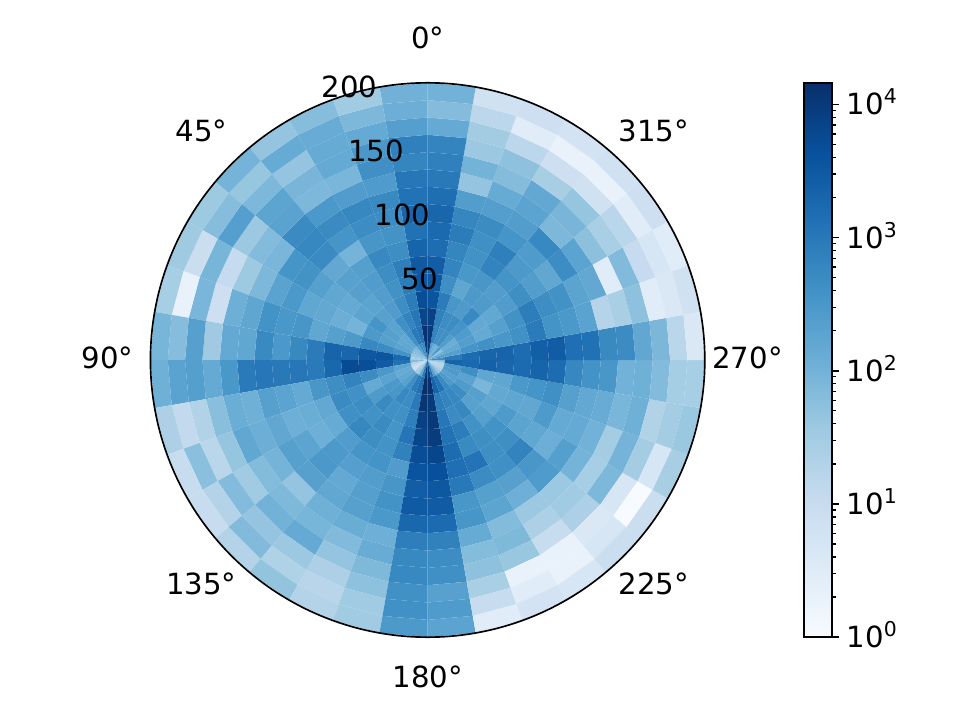}
    \caption{}
    \label{subfig:polar_density_map}
\end{subfigure}
\begin{subfigure}{.38\textwidth}
    \centering
    \includegraphics[width=\textwidth]{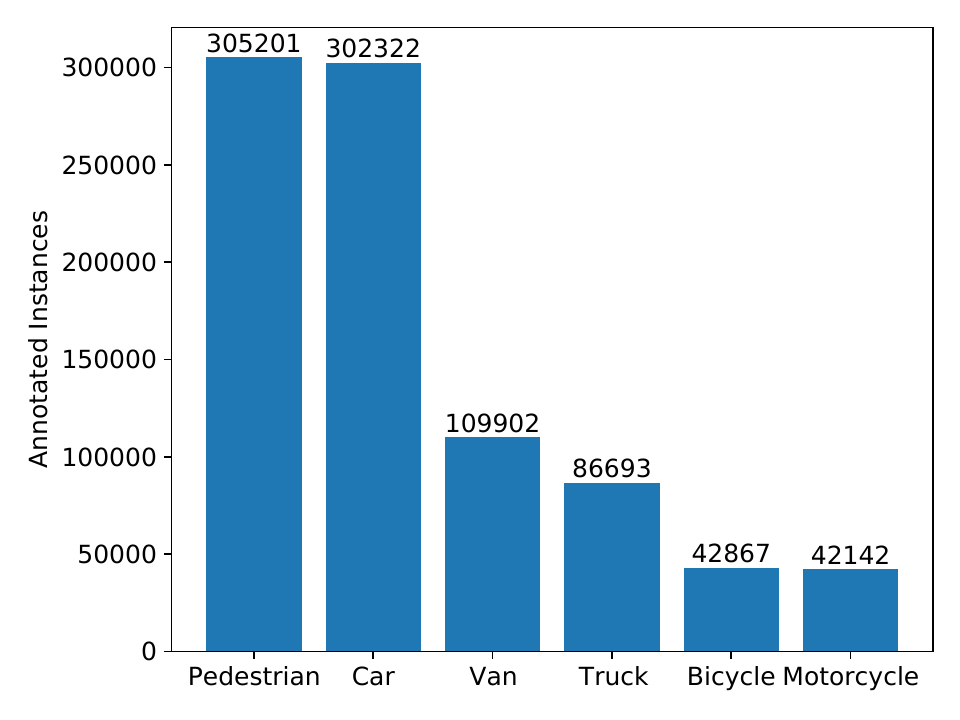}
    \caption{}
    \label{subfig:class_histogram}
\end{subfigure}
\caption{Statistics for Adver-City. (a) Percentage of frames pertaining to each weather condition. (b) Percentage of frames per time of day. (c) Number of objects within line of sight of ego vehicle per distance to ego on dense and sparse scenarios. (d) Number of keyframes per annotation counts. (e) Polar density map in log scale for objects within line of sight of ego vehicle. Distance (in meters) and angle are in relation to ego, with the scale shown to the right. (f) Number of annotated instances per object class. Objects that are simultaneously detected by multiple viewpoints are not counted multiple times.}
\label{fig:stats}
\end{figure*}

Figure~\ref{fig:stats} presents statistics for the Adver-City dataset. On average, scenarios last $21.9$ seconds ($219$ frames at $10$Hz) and contain $52$ unique vehicles and $44$ unique pedestrians. Overall, the dataset has $24,087$ frames and $890,127$ annotations pertaining to $11,418$ vehicles and $4,840$ pedestrians.

Figure~\ref{subfig:weather} and Figure~\ref{subfig:time_of_day} highlight the diversity of environmental conditions in Adver-City, showing slight variations in the percentage of frames for each weather condition. These differences result from running scenarios individually, leading to minor variations in agent behaviour and consequently in scenario length. Figure~\ref{subfig:density} shows the impact of density settings, with dense scenarios having significantly more objects, especially near the ego vehicle, compared to sparse ones.

Figure~\ref{subfig:annotations} and Figure~\ref{subfig:polar_density_map} showcase the diversity of frames in Adver-City. The first shows the balance between the number of annotations in the keyframes, ranging from just a few to over $120$ bounding boxes. The second shows that objects are positioned at various angles and distances from the ego vehicle, with a concentration around axial angles due to road layout. Figure~\ref{subfig:class_histogram} highlights that the Pedestrian category has the highest number of annotated instances, making Adver-City well-suited for detecting vulnerable road users.
\section{Benchmark}
\label{sec:benchmark}

How can a collaborative multi-modal 3D object detection model trained without adverse weather generalize to adverse weather conditions? To answer that question, we used the CoBEVT~\cite{xu_2022_cobevt} model trained on OPV2V~\cite{xu_2021_opv2v} (which has no adverse weather) to perform inference on Adver-City, having as a baseline the Culver City test scenario from OPV2V.

We used the trained model provided by the authors of HEAL~\cite{lu_2024_heal}, which has PointPillars~\cite{lang_2019_pointpillars} as the backbone for LiDAR, and Lift-Splat-Shoot~\cite{phillion_2020_lift} for camera with EfficientNet~\cite{tan_2019_efficientnet} as the image encoder.
 
Even though our dataset follows the OPV2V annotation schema, minor adjustments were required to ensure the OPV2V-trained model performed adequately on Adver-City. Since OPV2V includes only the `car' category, the model was limited to detecting cars. Bounding boxes for dissimilar categories such as `pedestrian', `bicycle', and `motorcycle', were removed from the experiment. Also, data from RSUs was excluded since OPV2V does not support V2X collaboration.

Then, to get a benchmark of the dataset, we took the same CoBEVT model (pre-trained on OPV2V) and trained it for 20 more epochs on Adver-City with a learning rate of $0.00001$. We split our dataset by road configurations, assigning the rural intersection, rural straight non-junction and urban non-junction scenarios to the training set, rural curved non-junction to the validation set, and urban intersection to the test set. As such, each set presents all weather conditions and density settings while avoiding data leakage. The resulting frame count for the train/validation/test split was $14,910/4,774/4,403$. Since further training was performed on the model, we included the data from the RSUs, but kept the `pedestrian', `bicycle', and `motorcycle' categories out of the experiment to guarantee a fair comparison with the OPV2V-trained model.

The results of both experiments are presented in Table~\ref{tab:benchmark}. For both, detection was evaluated for objects near the ego vehicle in a range of $x \in [-102.4, 102.4]$, $y \in [-102.4, 102.4]$, with CAVs having a communication range of $70$ meters. For the OPV2V-trained model, we performed inference on the entire Adver-City dataset, with the five road configurations described in Section~\ref{subsec:scenarios} present in each combination of adverse weather conditions and density settings.

\begin{table*}
\centering
\begin{tabular}{cc|clcl|clcl}
\toprule
\multirow{3}{*}{\textbf{Dataset}} & \multirow{3}{*}{\textbf{Scenario}} & \multicolumn{4}{c|}{\textbf{\begin{tabular}[c]{@{}c@{}}Trained on OPV2V\\ (LiDAR + Camera, AP@50)\end{tabular}}} & \multicolumn{4}{c}{\textbf{\begin{tabular}[c]{@{}c@{}}Trained on Adver-City\\ (LiDAR + Camera, AP@50)\end{tabular}}} \\ \cmidrule{3-10} 
&          & \multicolumn{2}{c}{Day}       & \multicolumn{2}{c|}{Night}   & \multicolumn{2}{c}{Day}         & \multicolumn{2}{c}{Night}      \\ \cmidrule{3-10} 
&          & Sparse            & \multicolumn{1}{c}{Dense}           & Sparse           & \multicolumn{1}{c|}{Dense}          & Sparse& \multicolumn{1}{c}{Dense}            & Sparse            & \multicolumn{1}{c}{Dense}            \\ \midrule
% \multirow{6}{*}{Adver-City}       & Clear Weather         & $46.70$           & $41.53$   & $42.76$          & $39.50$   &       &            &      &            \\
% & Soft Rain& $42.30$           & $40.45$   & $39.42$          & $34.78$   &       &            &      &            \\
% & Heavy Rain            & $43.06$           & $38.72$   & $34.11$          & $38.03$   &       &            &      &            \\
% & Fog      & $38.68$           & $36.68$   & $37.55$          & $34.11$   &       &            &      &            \\
% & Foggy Heavy Rain      & $37.83$           & $35.77$   & $40.40$          & $34.18$   &       &            &      &            \\
% & Glare    & $42.43$           & $38.55$   & \multicolumn{2}{c|}{-}       &       &            & \multicolumn{2}{c}{-}          \\ \midrule
\multirow{6}{*}{Adver-City}       & Clear Weather                      & $46.70$           & $41.53$ & $42.76$          & $39.50$ & $56.55$            & $47.88$  & $57.94$            & $51.83$ \\
& Soft Rain                          & $42.30$           & $40.45$ & $39.42$          & $34.78$ & $60.59$            & $52.70$  & $59.60$            & $49.07$ \\
& Heavy Rain                         & $43.06$           & $38.72$ & $34.11$          & $38.03$ & $63.18$            & $49.01$  & $56.49$            & $51.83$ \\
& Fog    & $38.68$           & $36.68$ & $37.55$          & $34.11$ & $59.82$            & $47.54$  & $59.06$            & $49.10$ \\
& Foggy Heavy Rain                   & $37.83$           & $35.77$ & $40.40$          & $34.18$ & $59.25$            & $48.80$  & $60.44$            & $50.59$ \\
& Glare  & $42.43$           & $38.55$ & \multicolumn{2}{c|}{-}     & $58.97$            & $48.32$  & \multicolumn{2}{c}{-}  \\ \midrule
OPV2V   & Culver City           & \multicolumn{2}{c}{$70.27$}   & \multicolumn{2}{c|}{-}       & \multicolumn{4}{c}{-}      \\ \bottomrule
\end{tabular}
\caption{Results for 3D object detection from the collaborative CoBEVT~\cite{xu_2022_cobevt} model. The model trained on OPV2V~\cite{xu_2021_opv2v} was used for inference on the entire Adver-City dataset to assess its robustness to challenging weather conditions, while the model trained on Adver-City was used only on its test set.}
\label{tab:benchmark}
\end{table*}

Regarding the results of the model trained on OPV2V, differences in scenario location, sensor configuration, and positioning caused a performance drop between the Culver City and Adver-City scenarios. While OPV2V was captured exclusively in urban scenarios with $64$-channel LiDARs and frontal cameras mounted on the vehicle's hood, our dataset was captured in both rural and urban scenarios with $32$-channel LiDARs and frontal cameras mounted on the windshield. 

As expected from a model trained without adverse weather, the best result occurred for the clear weather scenario during the day, with the lowest result for heavy rain during the night. Despite CARLA's limitations, performance dropped noticeably as weather conditions became more challenging, with a gap of $12.59$ between the highest and lowest scoring scenarios. Increasing object density from sparse to dense decreased performance for most scenarios, indicating the effects of occlusion on the model's detection.

When trained on Adver-City, however, the model's results showed a smaller gap to Culver City, indicating that it adequately learned from our dataset's distinct weather conditions and sensor configurations. Although the results for clear weather scenarios during the day improved, nighttime and adverse weather scenarios exhibited a significantly higher improvement, an effect of OPV2V's scenarios exclusively portraying clear weather during the day and illustrating the importance of utilizing adverse weather data for training models.

The detection scores of the CoBEVT model trained on Adver-City over the entire test set were $58.30/52.44/38.90$ (AP@30/50/70). In Table~\ref{tab:cobevt_results}, we compare it with the results of this model provided by~\cite{lu_2024_heal} on other datasets, highlighting the impact that adverse weather conditions have on collaborative perception models.

\begin{table}
\centering
\begin{tabular}{c|c}
\toprule
                    & \textbf{\begin{tabular}[c]{@{}c@{}}LiDAR + Camera\\AP@50\end{tabular}} \\ \midrule
\textbf{Adver-City} & 52.4                  \\
\textbf{DAIR-V2X}   & 63.8                  \\
\textbf{OPV2V}      & 88.5                  \\ \bottomrule
\end{tabular}
\caption{Results for 3D object detection for CoBEVT~\cite{xu_2022_cobevt} in distinct collaborative perception datasets. Results for DAIR-V2X~\cite{yu_2022_dairv2x} and OPV2V~\cite{xu_2021_opv2v} provided by \cite{lu_2024_heal}.}
\label{tab:cobevt_results}
\end{table}

\section{Limitations}
\label{sec:limitations}

One of the main benefits of simulation comes from the amount of control over scenarios. It enables uncommon real-life conditions such as heavy fog and intense glare to be seamless and extensively experimented with. Moreover, simulation is more scalable and accessible than real-world data acquisition, which requires costly sensor equipment to be captured. However, such benefits do not come without downsides, the main one being a reduction in realism. Despite recent advances, current simulators are still not capable of accurately simulating complex real-world phenomena such as adverse weather conditions.

For that matter, we found it imperative to assess the level of realism of Adver-City, comparing it to real-world datasets with extensive adverse weather samples. Due to the scarcity of adverse weather conditions in real-world CP datasets, we selected three single-vehicle datasets for this comparison. The first is BDD100K~\cite{yu_2020_bdd100k}, collected in the USA and featuring diverse weather conditions such as fog, rain, and snow, although only containing camera data. The second is NuScenes~\cite{caesar_2020_nuscenes}, one of the most used single-vehicle datasets to date, featuring multi-modal data collected in the USA and Singapore of clear weather and rainy scenes. Lastly, Boreas~\cite{burnett_2023_boreas} is a multi-modal Canadian dataset containing rainy, snowy, and sunny scenarios. 

We restricted our analysis to images of rainy scenarios during the day, since not all selected datasets have sufficient samples of other types of data such as LiDAR point clouds or foggy scenarios. Additionally, since all three single-vehicle datasets were collected in urban settings, we selected image samples from Adver-City's urban scenarios (urban intersection and urban non-junction) in rainy conditions (soft rain and heavy rain) collected by the frontal camera from the ego vehicle or the CAVs. This criterion restricted the analysis to $4,443$ images from Adver-City, for which an equal number of images matching the criterion were randomly selected from the real-world datasets.

Inspired by~\cite{reinhardt_2024_toward}, we evaluated the domain gap between synthetic and real data using t-SNE~\cite{maaten_2008_tsne} and FID (Fréchet Inception Distance)~\cite{heusel_2017_gans}. We used Scikit-Learn's~\cite{pedregosa_2011_scikit-learn} implementation of t-SNE calculating it from features computed by EfficientNet~\cite{tan_2019_efficientnet}, and~\cite{seitzer_2020_fid}'s implementation of FID.

\begin{figure}%[!hbt]
    \includegraphics[width=.474\textwidth]{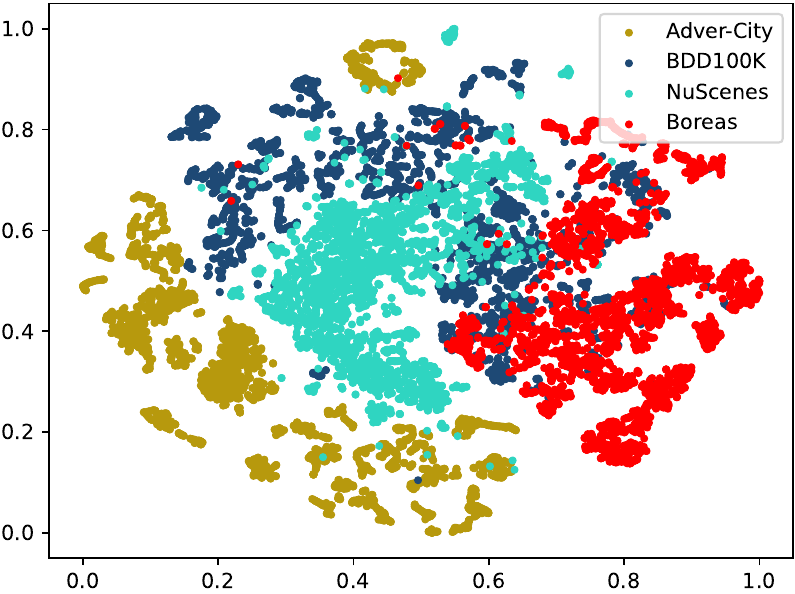}
    \caption{t-SNE plot generated with Scikit-Learn~\cite{pedregosa_2011_scikit-learn} from features computed by EfficientNet~\cite{tan_2019_efficientnet} of each dataset's rainy images. Adver-City's clusters have lower overlap than the real-world datasets, yet still having minor overlap with them.}
    \label{fig:tsne_plot}
\end{figure}

The t-SNE plot, presented in Figure~\ref{fig:tsne_plot}, illustrates the domain gap between each of the datasets analyzed, with similar samples grouped together. From the amount of overlap between samples in the plot, it can be noticed that samples from Adver-City have less in common with the real-world datasets than they have among themselves. Yet, many regions of the plot show overlap between Adver-City's samples and samples from each of the real-world datasets. For a better visualization of the samples and common characteristics among distinct regions of the plot, see the Supplementary Material for a t-SNE plot with points replaced by their corresponding images.

Even though the t-SNE plot shows that Adver-City has some level of similarity to the real-world datasets, it does not provide a quantitative evaluation of that similarity, for which the FID score brings a clearer metric. Table~\ref{tab:fid_rain} shows the FID scores for the rainy images of each pair of datasets, showing that Adver-City's FID score compared to both BDD100K ($147.1$) and NuScenes ($145.2$) is roughly $50\%$ greater than the score between the two of them ($95.6$). However, the BDD100K-Boreas's score ($189.5$) is almost double the value of the BDD100K-NuScenes score. As can be observed in the Supplementary Material, Boreas's distinctiveness can be attributed to the presence of multiple raindrops on the camera lens, a feature not observed in the other datasets. This distinctive feature makes Boreas more dissimilar to BDD100K and NuScenes than Adver-City, indicating that the synthetic dataset's realism is within an acceptable margin.

\begin{table}%[!htb]
\centering
\resizebox{0.478\textwidth}{!}{\begin{tabular}{c|cccc}
\toprule
           & \textbf{Adver-City} & \textbf{BDD100K} & \textbf{NuScenes} & \textbf{Boreas} \\ \midrule
\textbf{Adver-City} & $0$          & $147.1$  & $145.2$    & $205.4$   \\ 
\textbf{BDD100K}    & $147.1$      & $0$      & $95.6$     & $189.5$   \\ 
\textbf{NuScenes}   & $145.2$      & $95.6$   & $0$        & $160.7$   \\ 
\textbf{Boreas}     & $205.4$      & $189.5$  & $160.7$    & $0$       \\ \bottomrule
\end{tabular}}
\caption{FID scores~\cite{heusel_2017_gans} comparing rainy image samples between Adver-City and real-world single-vehicle datasets. A value of $0$ indicates an equal distribution, while higher values indicate more distinct features. BDD100K~\cite{yu_2020_bdd100k} and NuScenes~\cite{caesar_2020_nuscenes} have the most similar rainy samples, while Boreas~\cite{burnett_2023_boreas} has the most distinct rainy scenes of all datasets analyzed.}
\label{tab:fid_rain}
\end{table}

Another way to assess the realism of Adver-City's synthetic weather is to evaluate the FID score between its rainy and clear weather samples, then compare this score with that of real-world datasets. As presented in Table~\ref{tab:fid_clear_rain}, this FID score for Adver-City is smaller than that of real-world datasets, which also suggests a lower realism. However, it is merely $17\%$ smaller than BDD100K's score and $23\%$ smaller than NuScenes, indicating a fair level of realism.

\begin{table}%[!htb]
\centering
\begin{tabular}{c|c}
\toprule
                    & \textbf{Rain VS. Clear} \\ \midrule
\textbf{Adver-City} & 60.7                  \\
\textbf{BDD100K}    & 72.9                  \\
\textbf{NuScenes}   & 79.4                  \\
\textbf{Boreas}     & 184.8                 \\  \bottomrule
\end{tabular}
\caption{FID scores between each dataset's rainy and clear weather images. Adver-City's lower score indicates the difference between its rainy and clear weather scenes is smaller than the difference found in real-world datasets. However, its gap is merely $17\%$ smaller than BDD100K's gap.}
\label{tab:fid_clear_rain}
\end{table}

Thus, our analysis shows that Adver-City is less realistic than real-world datasets, an expected feature of a synthetic dataset. However, we have shown that it is not significantly dissimilar from them, making it a valuable and viable source of experimentation for CP models focused on adverse weather, especially considering that a real-world CP dataset focused on adverse weather does not yet exist.
\section{Conclusions}
\label{sec:conclusions}

We introduced Adver-City, the first open-source dataset for Collaborative Perception (CP) focused on adverse weather conditions. Its scenarios, based on real-world data on adverse weather and poor visibility conditions, provide a valuable testbed for experimentation. Additionally, its variations in scene density and weather offer unique challenges for models tackling CP in complex environments.

Our experiments showed that models trained without adverse weather scenes suffer a significant drop in performance when used for inference in these conditions, whereas models trained with them still face a significant challenge. In the future, we hope to investigate the robustness of CP models to adverse weather conditions and improve upon their designs to address shortcomings arising from their use in adverse weather.

As an open-source dataset, Adver-City's open code can be seamlessly extended to include additional weather conditions such as snow or dust, new road configurations, and maps that leverage weather and density variations. Sensor configurations can also be adjusted to generate new data or enhance the realism of CARLA's sensors. We hope that its open-source nature encourages users to share their work, extending the benefits of collaboration beyond the confines of perception algorithms.
{
    \small
    \bibliographystyle{ieeenat_fullname}
    \bibliography{bibliography}
}

\clearpage
\setcounter{page}{1}
\maketitlesupplementary

\section{Road Configurations}
\label{sec:road_configurations}

\begin{figure}%[!htb]
    \centering
    \includegraphics{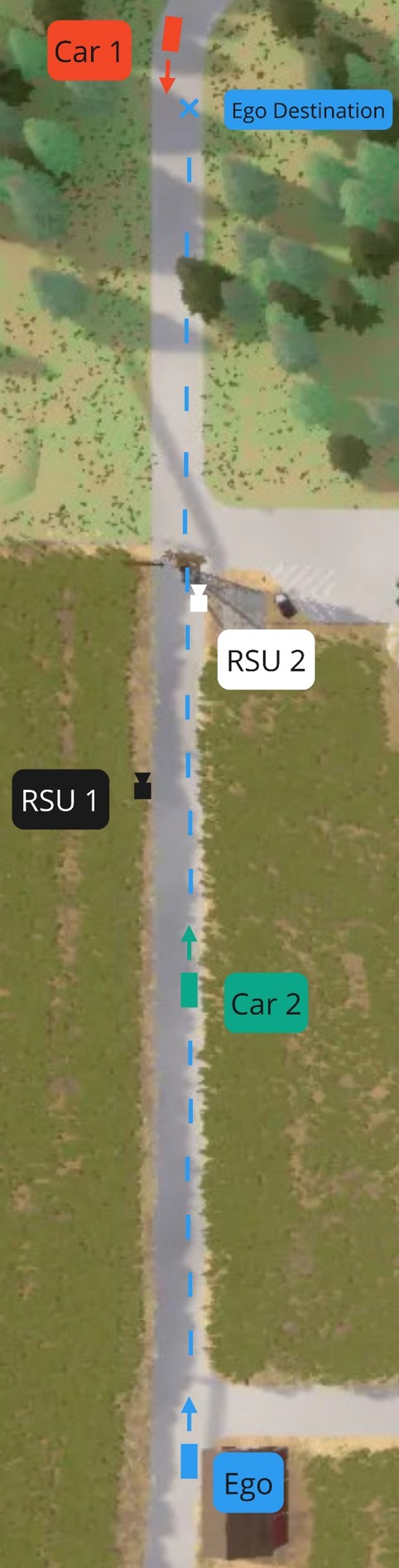}
    \caption{Rural straight non-junction configuration.}
    \label{fig:rsnj_layout}
\end{figure}

\begin{figure}%[!htb]
    \centering
    \includegraphics{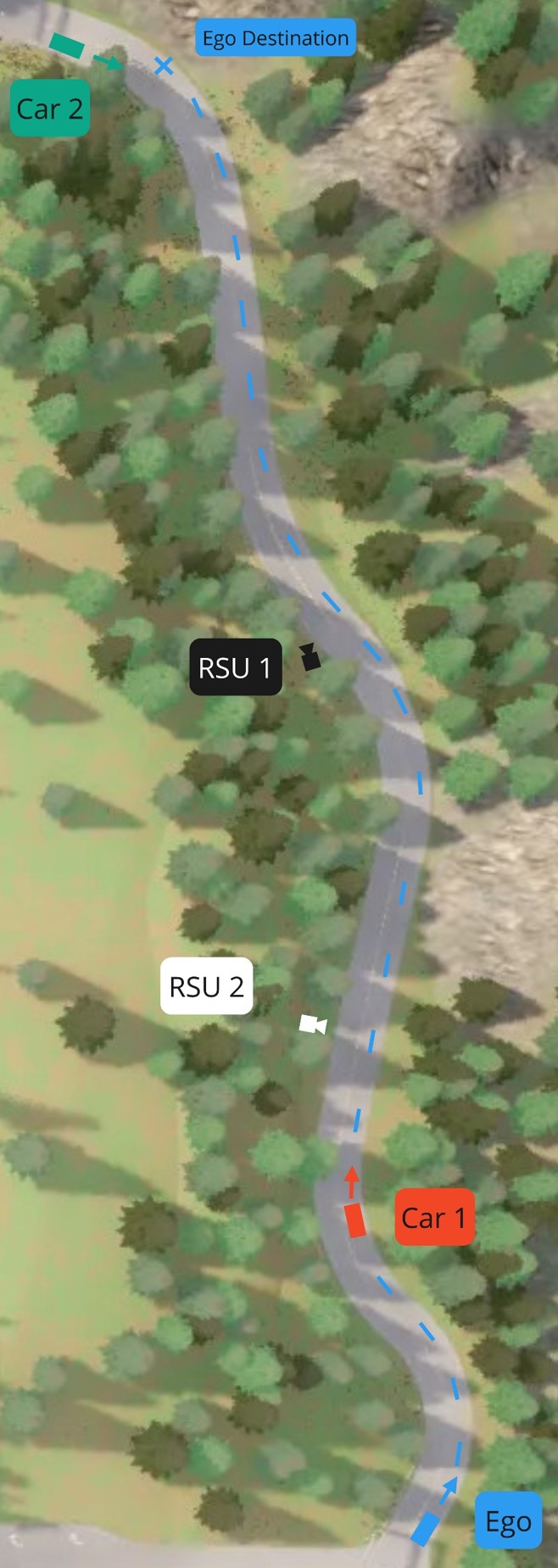}
    \caption{Rural curved non-junction configuration.}
    \label{fig:rcnj_layout}
\end{figure}

\begin{figure}%[!htb]
    \centering
    \includegraphics[width=.478\textwidth]{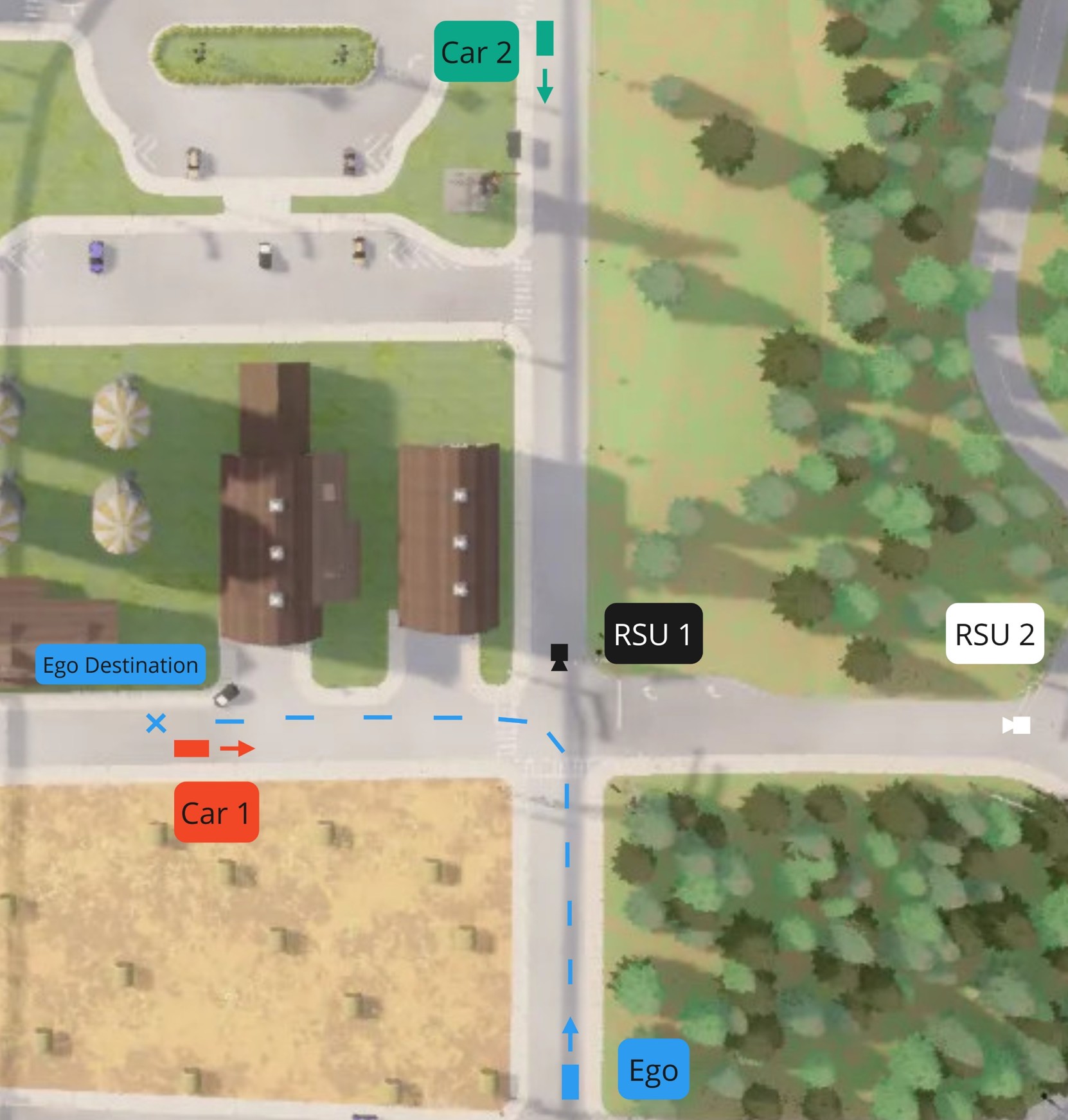}
    \caption{Rural intersection configuration.}
    \label{fig:ri_layout}
\end{figure}

\begin{figure}%[!htb]
    \centering
    \includegraphics[width=.478\textwidth]{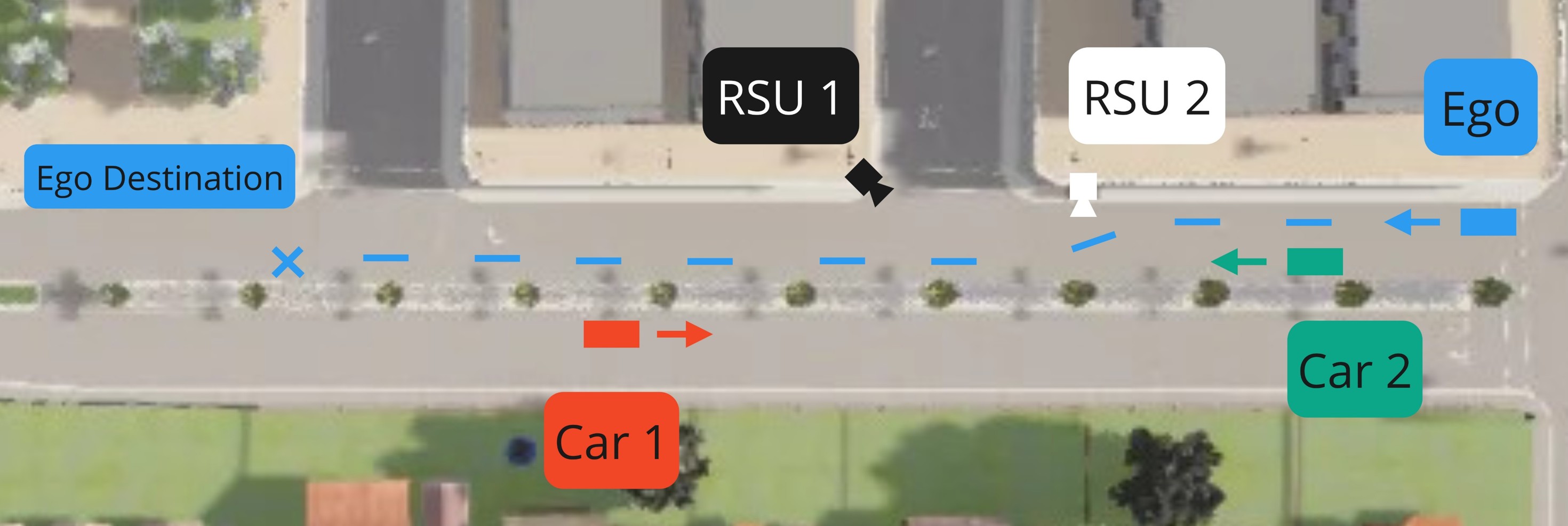}
    \caption{Urban non-junction configuration.}
    \label{fig:unj_layout}
\end{figure}

\begin{figure}%[!htb]
    \centering
    \includegraphics[width=.478\textwidth]{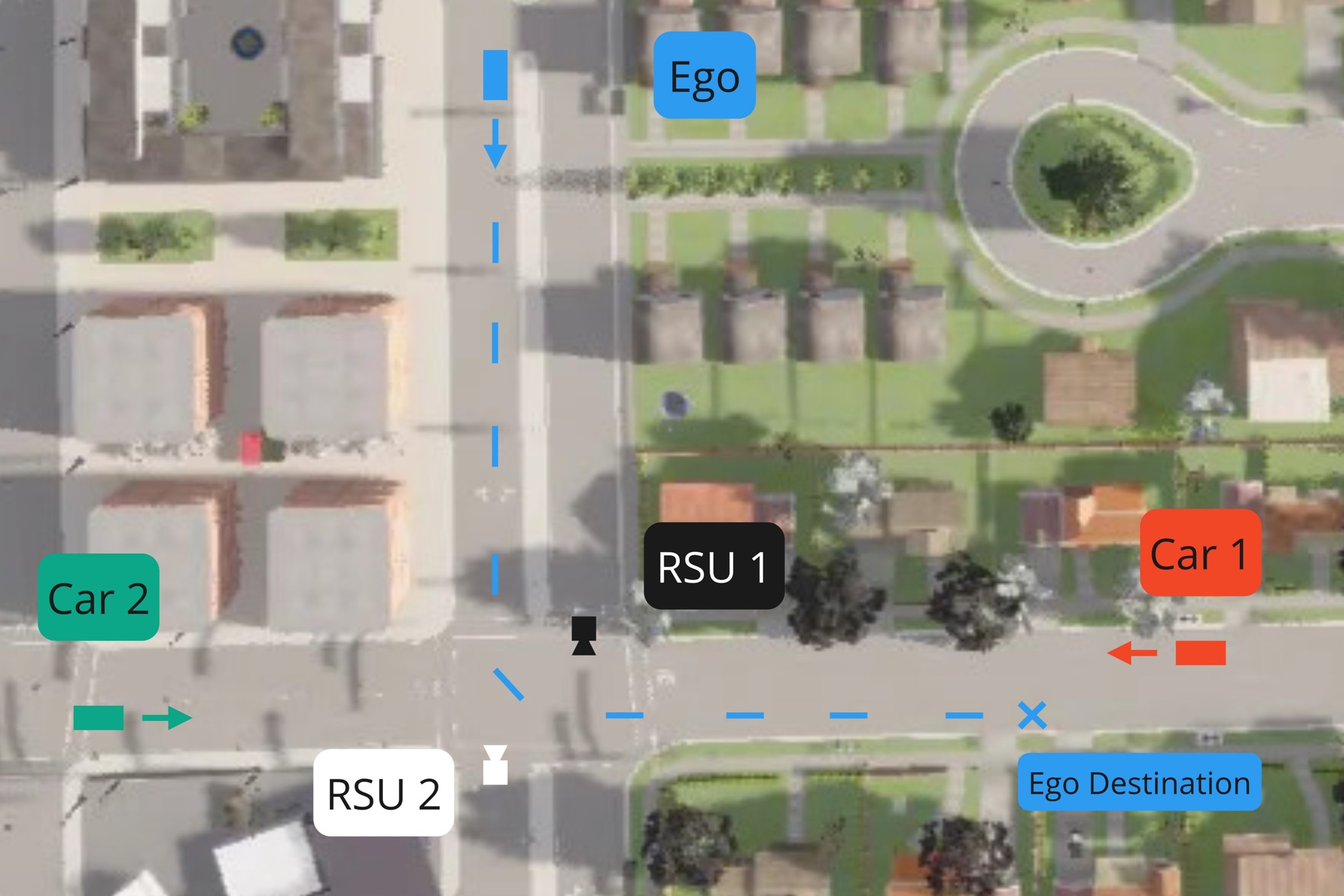}
    \caption{Urban intersection configuration.}
    \label{fig:ui_layout}
\end{figure}

The placement of CAVs (Connected Autonomous Vehicles) and RSUs (Roadside Units) for each road configuration used in Adver-City is presented in Figures \ref{fig:rsnj_layout}, \ref{fig:rcnj_layout}, \ref{fig:ri_layout}, \ref{fig:unj_layout}, and \ref{fig:ui_layout}.

The rural straight non-junction configuration (Figure~\ref{fig:rsnj_layout}) accounts for the longest path to be driven by the ego vehicle, creating the longest scenarios of the dataset. The ego's path goes through a yielding intersection, then to a straight surrounded by crops on both sides, reaching an intersection with a stop sign and then proceeding to its destination. Car~1 starts closer to the ego's destination but proceeds towards the ego's starting position, while Car~2 starts ahead of the ego and goes to the same direction as it. RSU~1 is positioned on top of a Stop sign, while RSU~2 is positioned on a windmill tower at a higher elevation.

In the rural curved non-junction configuration (Figure~\ref{fig:rcnj_layout}), the ego vehicle goes through a curved road layout that first goes uphill and then downhill and is surrounded by trees on both sides. Car~1 begins the scenario ahead of the ego, going towards the same destination, while Car~2 starts closer to the ego's destination and proceeds towards the ego's starting position. Both RSUs are placed next to the road at ground height, simulating mobile RSUs commonly used by law enforcement.

The rural intersection configuration (Figure~\ref{fig:ri_layout}) has the shortest route to be taken by the ego vehicle, in which it approaches the intersection through a short straight with trees to its right and an open field to its left, waits for the traffic light to turn green, and then makes a left turn, briefly going through a section with buildings to its right and the open field to the left. Car~1 approaches the intersection from the West, while Car~2 approaches it from the North, both of them waiting for it to turn green until the end of the scenario. RSU~1 is positioned on top of the traffic light, while RSU~2 is positioned on top of an adjacent intersection's traffic light, providing a unique viewpoint to the agents.

In the urban non-junction configuration (Figure~\ref{fig:unj_layout}), the ego vehicle follows a straight road with a median strip, with tall walls to the right and houses to the left, performing a lane change shortly after the start of the scenario and then following a straight line until reaching its destination. Car~1 is positioned on the opposing traffic lane, moving towards the ego's initial position, while Car~2 is slightly ahead of the ego, moving in the same direction as it, but starting the scenario on the lane immediately to the left of the ego. RSU~1 is positioned atop the tall wall to the right of the road, viewing the non-junction at an angle, while RSU~2 is positioned on top of a bus stop located at the side of the road.

The urban intersection configuration (Figure~\ref{fig:ui_layout}) is the one with the greatest number of objects, having the ego approach an intersection through a short two-way road, waiting for the traffic light to turn green, making a left turn towards a residential road and proceeding briefly through it. Car~1 is positioned on the residential road, approaching the intersection from the East, and Car~2 is positioned on the road West of the intersection, approaching it from that direction, both of them waiting by the intersection for the traffic light to turn green until the end of the scenario. Both RSUs are positioned on  traffic lights at the intersection, offering a surrounding view of it.

\section{Comparison with Real-Life Datasets}
\label{sec:comparison}

\begin{figure*}%[!htb]
    \includegraphics[width=\textwidth]{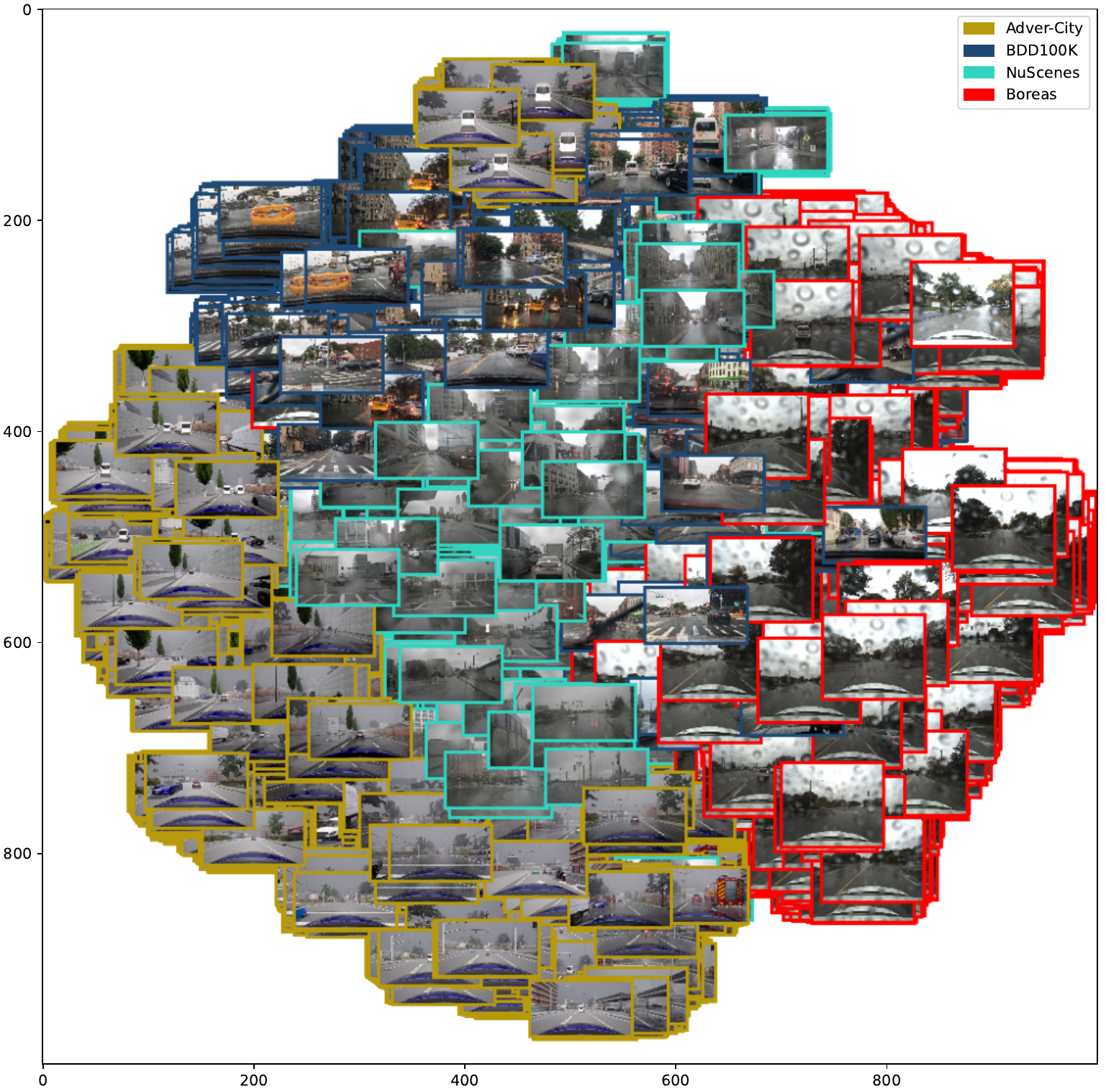}
    \caption{t-SNE plot with points replaced by their corresponding dataset samples.}
    \label{fig:tsne_samples}
\end{figure*}

Based on the t-SNE plot presented in Figure~\ref{fig:tsne_plot}, we replaced each point with their corresponding sample to better visualize and interpret the data, with the new plot presented in Figure~\ref{fig:tsne_samples}.

A distinguishing characteristic of the Boreas dataset~\cite{burnett_2023_boreas}, based on the plot, is that most samples have rain splatter on the camera lens, a feature that can only be observed in a few samples of the NuScenes dataset~\cite{caesar_2020_nuscenes}.

At the top of the plot, a small cluster of Adver-City samples was separated from the rest of the dataset due to its similarity to samples from the BDD100K dataset~\cite{yu_2020_bdd100k}, this being the presence of a white van in front of the ego vehicle.

Looking at the main Adver-City cluster, it is possible to note similar characteristics between its samples and those of the other datasets that were positioned adjacent to it. NuScenes samples positioned closer to our dataset's main cluster have as a distinguishing characteristic a cloudy gray sky similar to those of the simulated dataset. Samples from the BDD100K dataset with high vehicle density were positioned to the left of the plot closer to samples from our dataset that also feature high vehicle density. Then, looking at the vicinity between the Adver-City and the Boreas clusters, we can observe that all the samples are characterized by having a wet pavement.

\end{document}